\documentclass[10pt,twocolumn,letterpaper]{article}

\usepackage{multirow}
\usepackage{tablefootnote}
\usepackage{colortbl}

\usepackage[ruled,linesnumbered]{algorithm2e}
\usepackage[pagenumbers]{cvpr} % To force page numbers, e.g. for an arXiv version

\usepackage[dvipsnames]{xcolor}

\definecolor{cvprblue}{rgb}{0.21,0.49,0.74}
\usepackage[pagebackref,breaklinks,colorlinks,citecolor=cvprblue]{hyperref}

\def\paperID{3614} % *** Enter the Paper ID here
\def\confName{CVPR}
\def\confYear{2024}

\title{Theoretically Achieving Continuous Representation of Oriented Bounding Boxes}

\author{\textbf{Zikai Xiao$^{1}$}, \textbf{Guoye Yang$^{1}$}, \textbf{Xue Yang$^{2}$}, \textbf{Taijiang Mu$^{1}$}\thanks{Corresponding author.} , \textbf{Junchi Yan$^{2}$}, \textbf{Shimin Hu$^{1}$}\\
{$^{1}$BNRist, Department of Computer Science and
Technology, Tsinghua University}\\ 
{$^{2}$Department of CSE \& MoE Key Lab of AI, Shanghai Jiao Tong University}\\
{\tt\small \{xzk23, yanggy19\}@mails.tsinghua.edu.cn \quad \{taijiang, shimin\}@tsinghua.edu.cn }\\
{\tt\small \{yangxue-2019-sjtu, yanjunchi\}@sjtu.edu.cn}\\
{\tt\small Code: \url{https://github.com/514flowey/JDet-COBB}}
}

\begin{document}
\maketitle
\begin{abstract}
Considerable efforts have been devoted to Oriented Object Detection (OOD). However, one lasting issue regarding the discontinuity in Oriented Bounding Box (OBB) representation remains unresolved, which is an inherent bottleneck for extant OOD methods. This paper endeavors to completely solve this issue in a theoretically guaranteed manner and puts an end to the ad-hoc efforts in this direction. Prior studies typically can only address one of the two cases of discontinuity: rotation and aspect ratio, and often inadvertently introduce decoding discontinuity, e.g. Decoding Incompleteness (DI) and Decoding Ambiguity (DA) as discussed in literature. Specifically, we propose a novel representation method called \textbf{C}ontinuous \textbf{OBB} (\textbf{COBB}), which can be readily integrated into existing detectors e.g. Faster-RCNN as a plugin. It can theoretically ensure continuity in bounding box regression which to our best knowledge, has not been achieved in literature for rectangle-based object representation. For fairness and transparency of experiments, we have developed a modularized benchmark based on the open-source deep learning framework Jittor's detection toolbox JDet for OOD evaluation. On the popular DOTA dataset, by integrating Faster-RCNN as the same baseline model, our new method outperforms the peer method Gliding Vertex by 1.13\% mAP\textsubscript{50} (relative improvement 1.54\%), and 2.46\% mAP\textsubscript{75} (relative improvement 5.91\%), without any tricks.

\end{abstract}

\begin{figure}[t!]
    \centering
    \subfloat[Rotation Discontinuity]{
        \label{fig:rotation_discontinuity}
        \begin{minipage}[b]{.23\textwidth}
            \centering
            \includegraphics[width=0.9\linewidth]{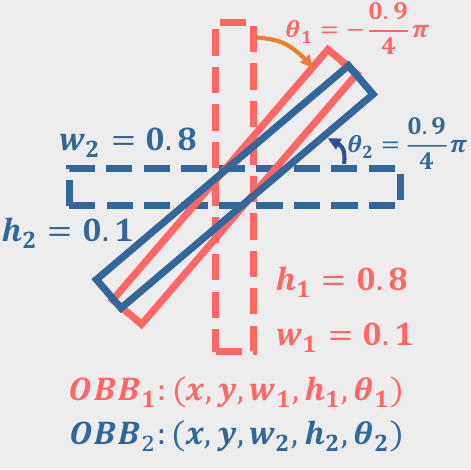}		
        \end{minipage}%
    }%
    \subfloat[Aspect Ratio Discontinuity]{
        \label{fig:aspect_ratio_discontinuity}
        \begin{minipage}[b]{.23\textwidth}
            \centering
            \includegraphics[width=0.9\linewidth]{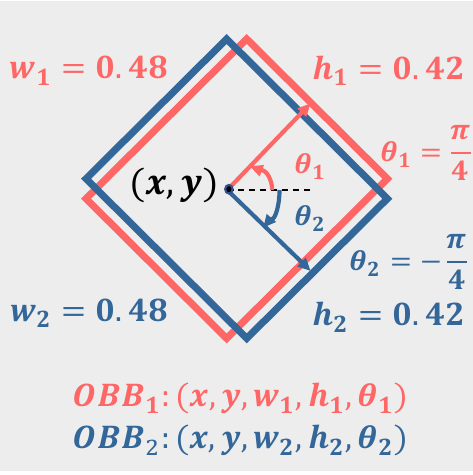}		
        \end{minipage}%
    }\\
    \subfloat[Decoding Incompleteness]{
        \label{fig:DI}
        \begin{minipage}[b]{.23\textwidth}
            \centering
            \includegraphics[width=0.9\linewidth]{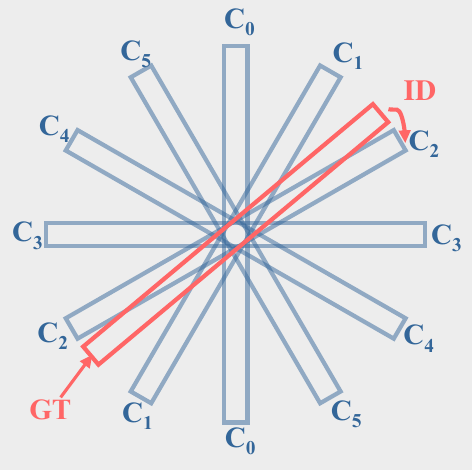}		
        \end{minipage}%
    }%
    \subfloat[Decoding Ambiguity]{
        \label{fig:DA}
        \begin{minipage}[b]{.23\textwidth}
            \centering
            \includegraphics[width=0.9\linewidth]{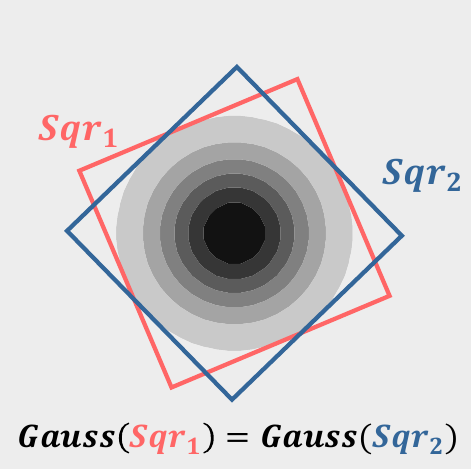}		
        \end{minipage}%
    }%
    \vspace{-1mm}
    \caption{\textbf{Examples of Discontinuity in OBB Representations.} 
    (a) Acute-angle Representation limits the rotation angle of OBBs inside a range of $\frac{\pi}{2}$ ($[-\frac{\pi}{4},\frac{\pi}{4})$ in this example). The red $\text{OBB}_1$ and the blue $\text{OBB}_2$ are similar, but their representations are significantly different.
    (b) Long-edge Representation determines the rotation angle $\theta$ by the long side and the x-axis. A slight disturbance in the aspect ratio of square-like OBBs will cause a huge change in their representation, which causes Aspect Ratio Discontinuity. (c) CSL~\cite{yang2020arbitrary} divides the rotation angle into several classifications (6 classifications in this figure). OBB between two classifications cannot be accurately represented, which brings DI. (d) GWD~\cite{yang2021rethinking} denotes OBBs by Gaussian distribution. As the squares with different rotation angles can correspond to the same Gaussian, the orientation of decoded squares will be ambiguous.}
    \label{fig:metrics}
    \vspace{-15pt}
\end{figure}

\section{Introduction}
Object detection constitutes a fundamental task within the realm of computer vision. In conventional object detection scenarios~\cite{8627998}, the commonplace approach involves the localization of objects using Horizontal Bounding Boxes (HBB). However, in many real-world settings such as remote sensing~\cite{xia2018dota, ding2021object} and scene text~\cite{8323240, Liao_2018_CVPR}, where objects exhibit arbitrary orientations, HBBs are inadequate in precisely delineating object boundaries. To overcome this issue, Oriented Bounding Boxes (OBB)~\cite{xia2018dota}, conceptualized as rotated rectangles, have been introduced as a more suitable representation for Oriented Object Detection (OOD).

Various models have been proposed for OOD~\cite{ding2019learning, ma2018arbitrary,
liu2017rotated, han2021redet, xu2020gliding}. However, as illustrated in Figs.~\ref{fig:rotation_discontinuity}-\ref{fig:aspect_ratio_discontinuity}, prevalent representation methods for OBBs exhibit discontinuity issues, encoding similar OBBs into distinct vectors. This introduces challenges in training neural networks as regression targets for similar input features may differ significantly, potentially causing confusion and hindering the training process.
The relationship between two OBBs can be conceptualized as one being transformed into the other through geometric operations: translation, rotation, scaling, and aspect ratio changes. While translation and scaling are relatively benign, rotation and aspect ratio changes are primary sources of discontinuity.

Rotation discontinuity, often referred to as the ``Boundary Problem''~\cite{yang2020arbitrary} or ``Rotation Sensitive Error''~\cite{qian2021learning}, stems from the periodicity of rotation angles. Although prior efforts~\cite{yang2020arbitrary, yang2021dense, yu2023phase} have addressed this, most of them still suffer from discontinuity arising from changes in aspect ratio. Some other methods, \eg Gliding Vertex~\cite{xu2020gliding}, effectively address aspect ratio discontinuity. Nonetheless, these techniques continue to face challenges in overcoming rotation discontinuity. The discontinuity phenomena emerge during the encoding of OBBs into the regression target. Consequently, they can be termed encoding discontinuity.

Additionally, existing methods aim at resolving the encoding discontinuity issue, yet meanwhile would often bring about decoding discontinuity, namely Decoding Incompleteness (DI) and Decoding Ambiguity (DA). DI arises when OBBs cannot be accurately represented, often attributed to angle discretization and classification as exemplified by CSL~\cite{yang2020arbitrary} in Fig.~\ref{fig:DI}. DA, on the other hand, pertains to instances where distinct OBBs share similar representations, rendering predicted OBBs sensitive to minor disturbances in the model's output, exemplified by GWD~\cite{yang2021rethinking} in Fig.~\ref{fig:DA}. Fundamentally, DI and DA result in decoded OBBs lacking continuity concerning their representation. Hence, we categorize DI and DA as decoding discontinuities. The decoding discontinuities bring precision errors and directly degrade the prediction precision.

Due to the absence of a rigorous definition, prior approaches have often addressed discontinuity issues incompletely. To address this, we introduce formal continuity metrics, evaluating previous methods using these benchmarks. As a comprehensive solution to discontinuity problems, we propose \textbf{C}ontinuous \textbf{OBB} (\textbf{COBB})—a novel, continuous OBB representation satisfying all defined metrics. 
COBB employs nine parameters derived from continuous functions based on the outer Horizontal Bounding Box (HBB) and OBB area. This ensures continuity as the outer HBB and OBB area undergo continuous changes during shape transformations.
Our COBB can be easily integrated into existing OOD methods by simply replacing their original representations of OBB with ours.

We have developed a benchmark using the detection toolbox JDet of Jittor~\cite{hu2020jittor} which is an open-source deep learning framework friendly to vision tasks. In particular, a fair comparison across different models is made by aligning the data augmentation schemes and diverse techniques.

Our experiments on this benchmark demonstrate the effectiveness
of COBB across diverse datasets and baseline detectors, particularly demonstrating advantages in high-precision object detection. Notably, it achieves a 3.95\% improvement in $\text{mAP}_{75}$ when applied to Faster R-CNN on the DOTA Dataset. Detailed results are presented in Sec.~\ref{sec:experiments}.

Our contributions encompass the following aspects:
\begin{itemize}
    \item We systematically analyze the inherent discontinuity issues in existing OBB representation methods for OOD, and introduce formal metrics to assess their continuity.
    \item Building upon our findings, we introduce COBB, a fully continuous representation of OBBs.
    \item We construct a new benchmark for fair comparisons among OOD methods. Experiments on this benchmark validate the effectiveness of our approach, highlighting its advantages in high-precision OOD.
\end{itemize}
\section{Related Work}
\label{sec:related}
\subsection{Oriented Object Detection}
With the increasing adoption of deep learning in computer vision, object detection models~\cite{liu2016ssd, redmon2016you, wang2022pvt, zhou2021rgb, fu2022light, guo2022attention, lan2022arm3d} have emerged to enhance computers' capacity for recognizing objects in natural images.  Typically tailored for predicting HBBs, these models serve as the foundation for OOD when augmented with modules for OBB prediction. Rotated Faster R-CNN~\cite{ren2015faster} stands as a prominent baseline for OOD, replacing HBB regression targets with OBBs. Several OOD methods, such as RoI Transformer~\cite{ding2019learning}, Gliding Vertex~\cite{xu2020gliding}, ReDet~\cite{han2021redet}, and Oriented R-CNN~\cite{xie2021oriented}, follow a similar structure and can be implemented on the Rotated Faster R-CNN framework.

While many OOD models share structural similarities, detailed implementation differences exist (\eg Gliding Vertex~\cite{xu2020gliding} using ResNet101 as the backbone network, whereas CSL~\cite{yang2020arbitrary} employs ResNet50). To facilitate fair comparisons, we established a uniform pipeline with modular alternatives for implementing these models, minimizing implementation disparities.

\subsection{Discontinuity in Oriented Object Detection}
Methods aiming to handle the discontinuous representation of OBBs fall into three categories: Loss Improvement, Angle Encoding, and New OBB representation.

\textbf{Loss Imporvement.}
Modifying the loss is a direct way to mitigate sudden changes in loss values caused by encoding discontinuity. Approaches like RIL~\cite{ming2021optimization} and RSDet~\cite{qian2021learning} propose loss functions that approach zero as the model's output converges to various representations of the ground truth OBB. PIoU~\cite{chen2020piou} and SCRDet~\cite{yang2019scrdet} incorporate Intersection over Union (IoU) between prediction results and regression targets in their loss. GWD~\cite{yang2021rethinking}, KLD~\cite{yang2021learning}, and KFIoU~\cite{yang2022kfiou} convert OBBs into Gaussian distributions for IoU calculation, introducing potential DA for square-like objects. While showing empirical effectiveness in reducing the impact of discontinuity, these approaches do not provide a theoretical resolution to the problem.

\textbf{Angle Encoding.}
Several methods focus on addressing the Periodicity of Angular (PoA), a primary cause of encoding discontinuity~\cite{yang2020arbitrary}. CSL~\cite{yang2020arbitrary} discretizes the rotation angle into a heavy regression target, with subsequent improvements by DCL~\cite{yang2021dense}, GF\_CSL~\cite{wang2022gaussian}, MGAR~\cite{wang2022multigrained}, and AR-CSL~\cite{zeng2023ars}. While these methods enhance rotation continuity, most of them struggle with square-like objects and may introduce DI. PSC~\cite{yu2023phase}, FSTC~\cite{zhang2023trigonometric}, and ACM~\cite{xu2023rethinking} encode the rotation angle into a continuous vector, yet they still exhibit discontinuity for square-like objects.

\textbf{New OBB Representation.}
Other approaches explore alternative representations for OBBs instead of rectangles and rotation angles. Gliding Vertex~\cite{xu2020gliding} slides the four vertices of a HBB to construct an OBB. O\textsuperscript{2}D-Net~\cite{wei2020oriented} and BBAVectors~\cite{yi2021oriented} represent an OBB using its center point and vectors from the center point to midpoints of its sides. PolarDet~\cite{zhao2021polardet} and CRB~\cite{yuan2022feature} leverage polar coordinates, yet the rotation discontinuity still exists. DHRec~\cite{nie2022multi} represents OBBs with double horizontal rectangles but struggles with distinguishing symmetrical tilted thin OBBs.

To the best of our knowledge, no method achieves perfect elimination of discontinuity. Previous approaches either fail in specific boundary situations or introduce DI and DA. The proposed COBB in this paper provides the first completely continuous representation of OBBs.

\section{Theoretically Continuous Representation}
\label{sec:method}
In this section, we first introduce our devised metrics to assess the continuity of existing methods in Sec.~\ref{sec:bdd_metric}. While Sec.~\ref{sec:method_details} unveils our COBB that theoretically ensures continuity under these metrics. The continuity of COBB is rigorously demonstrated in Sec.~\ref{sec:continuity}, with comprehensive details provided in the supplemental material.

\begin{table}[t!]
	\renewcommand\arraystretch{1.4}
	\setlength\tabcolsep{4pt}
	\footnotesize
	\centering
	\caption{\textbf{Comparison of Methods Dedicated to Discontinuity}. 
Tar (R), Tar (A), Loss (R), and Loss (A) stand for Target Rotation Continuity, Target Aspect Ratio Continuity, Loss Rotation Continuity, and Loss Aspect Ratio Continuity, respectively. Dec (C) and Dec (R) stand for Decoding Completeness and Decoding Robustness. Acute-Angle is the common OBB representation that limits the rotation angle into a range of $\frac{\pi}{2}$ ($[-\frac{\pi}{4}, \frac{\pi}{4})$ for example). Long-Edge is another common OBB representation, which determines the rotation angle $\theta$ by the long side and the $x$-axis, and the $\theta$ is within a range of length $\pi$. Further explanation of this table is provided in the supplemental material, see Sec. \ref{sec:futher_explain}.}
\vspace{-5pt}
    \resizebox{0.48\textwidth}{!}
{
	\begin{tabular}{ccccccc}
\toprule
\textbf{Method} & \textbf{Tar (R)} & \textbf{Tar (A)} & \textbf{Loss (R)} & \textbf{Loss (A)} & \textbf{Dec (C)} & \textbf{Dec (R)} \\
\hline
Acute-Angle & - & \checkmark & - & \checkmark & \checkmark & \checkmark \\
Long-Edge & - & - & - & - & \checkmark & \checkmark \\\hline
RIL~\cite{ming2021optimization} & - & - & \checkmark & \checkmark & \checkmark & \checkmark \\
RSDet ($l_{mr}^{5p}$)~\cite{qian2021learning} & - & - & \checkmark & \checkmark & \checkmark & \checkmark \\
PIoU~\cite{chen2020piou} & - & - & \checkmark & \checkmark & \checkmark & \checkmark \\
SCRDet~\cite{yang2019scrdet} & - & - & \checkmark & \checkmark & \checkmark & \checkmark \\
GWD~\cite{yang2021rethinking} & \checkmark & \checkmark & \checkmark & \checkmark & \checkmark & - \\
KLD~\cite{yang2021learning} & \checkmark & \checkmark & \checkmark & \checkmark & \checkmark & - \\
KFIoU~\cite{yang2022kfiou} & \checkmark & \checkmark & \checkmark & \checkmark & \checkmark & - \\
\hline
CSL~\cite{yang2020arbitrary} & \checkmark & - & \checkmark & - & - & - \\
DCL~\cite{yang2021dense} & \checkmark & - & \checkmark & \checkmark & - & - \\
GF\_CSL~\cite{wang2022gaussian} & \checkmark & \checkmark & \checkmark & \checkmark & - & - \\
MGAR~\cite{wang2022multigrained} & - & - & \checkmark & - & \checkmark & \checkmark\\
AR-CSL~\cite{zeng2023ars} & \checkmark & - & \checkmark & - & - & -\\
PSC~\cite{yu2023phase} & \checkmark & - & \checkmark & - & \checkmark & \checkmark\\
FSTC~\cite{zhang2023trigonometric} & \checkmark & - & \checkmark & - & \checkmark & \checkmark \\
ACM~\cite{xu2023rethinking} & \checkmark & - & \checkmark & - & \checkmark & \checkmark \\
\hline
Gliding Vertex~\cite{xu2020gliding} & - & \checkmark & - & \checkmark & \checkmark & - \\
O\textsuperscript{2}D-Net~\cite{wei2020oriented} & - & \checkmark & - & \checkmark & \checkmark & \checkmark \\
BBAVectors~\cite{yi2021oriented} & - & \checkmark & - & \checkmark & \checkmark & - \\ 
PolarDet~\cite{zhao2021polardet} & - & \checkmark & - & \checkmark & \checkmark & \checkmark \\
DHRec~\cite{nie2022multi} & \checkmark & \checkmark & \checkmark & \checkmark & \checkmark & - \\
CRB~\cite{yuan2022feature} & - & - & \checkmark & - & \checkmark & \checkmark\\
\rowcolor{gray!20} Ours & \checkmark & \checkmark & \checkmark & \checkmark & \checkmark & \checkmark\\
\bottomrule
	\end{tabular}
 }
	\vspace{-8pt}
\label{tab:boundary_methods}
\end{table}
\subsection{Metrics for Continuity}
\label{sec:bdd_metric}

As depicted in Fig.~\ref{fig:metrics}, prevalent methods for OBB prediction commonly face the challenge of encoding discontinuity. Prior endeavors often address some specific boundary cases, such as nearly horizontal OBBs and square-like OBBs~\cite{xu2020gliding, yu2023phase, yang2020arbitrary}. While these methods may exhibit continuity in certain boundary scenarios, they often overlook others. For instance, CSL~\cite{yang2020arbitrary} maintains rotation continuity for nearly horizontal OBBs but fails for square-like OBBs. Furthermore, some methods assert the resolution of the discontinuity, yet they still struggle with sudden changes in regression targets~\cite{yu2023phase, xu2023rethinking}.

To formally define continuity, we introduce $f_{Enc}$ as the mapping function from an OBB to $R^n$, and $f_{Dec}$ as the reverse mapping from a subset of $R^n$ to an OBB. Notably, $R(x, \theta)$ denotes the transformation generating an OBB $y$ by rotating the initial OBB $x$ by $\theta$ in a clockwise direction. Meanwhile, $A(x, r)$ generates a set of OBBs $\{y, z\}$ by adjusting one side of OBB $x$ to be $r$ times its original length. We refer to $S$ as the set of OBBs, and $L$ symbolizes the loss function. Models utilize $f_{Enc}$ to convert OBBs into regression targets and employ $f_{Dec}$ to translate prediction results into estimated OBBs.

\textbf{Target Rotation Continuity:}
Minor rotations should minimally affect the regression target.
\begin{equation}
\small
    \forall x\in S, \lim_{\theta\to 0}||f_{Enc}(x)-f_{Enc}\left(R(x, \theta)\right)||=0.
\end{equation}
The corresponding discontinuity is often referred to as the ``Boundary Problem''~\cite{yang2020arbitrary, yang2021dense}.

\textbf{Target Aspect Ratio Continuity:} Slight changes in aspect ratio should minimally impact the regression target.
\begin{equation}
\small
    \forall x\in S, \lim_{r\to 1}\sum_{y\in A(x,r)}||f_{Enc}(x)-f_{Enc}(y)||=0.
\end{equation}

\textbf{Loss Rotation Continuity:}
Small rotations should minimally affect the loss value.
\begin{equation}
\small
    \forall x\in S, \lim_{\theta\to 0}||L\left(f_{Enc}(x),f_{Enc}\left(R(x, \theta)\right)\right)||=0.
\end{equation}

\textbf{Loss Aspect Ratio Continuity:}
Minor aspect ratio changes should minimally alter the loss value.
\begin{equation}
\small
    \forall x\in S, \lim_{r\to 1}\sum_{y\in A(x,r)}||L\left(f_{Enc}(x),f_{Enc}(y)\right)||=0.
\end{equation}

\textbf{Decoding Completeness:}
Every OBB can be accurately represented. 
\begin{equation}
\small
    \forall x\in S, \exists t \in R^n, IoU\left(x, f_{Dec}(t)\right) = 1.
\end{equation}
Decoding Completeness is equivalent to avoiding Decoding Incompleteness (DI) illustrated in Fig.~\ref{fig:DI}.

\textbf{Decoding Robustness:} Decoded OBBs should be robust to slight errors in their representation.
\begin{equation}
\small
\begin{aligned}
&\forall x\in S, \forall \epsilon>0, \exists \xi>0, \forall \Delta d\in R^n\land ||\Delta d||<\xi,
\\
&1-IoU\left(x, f_{Dec}\left(f_{Enc}(x)+\Delta d\right)\right)<\epsilon.
\end{aligned}
\end{equation}
Decoding Robustness is equivalent to avoiding Decoding Ambiguity (DA) illustrated in Fig.~\ref{fig:DA}. 

Previous research has covered the first four metrics, which we formally defined, whereas there's been limited exploration of the last two metrics. Our investigation into existing OBB representation methods helped unveil the neglected discontinuity, which formed the basis for these two metrics. A further detailed explanation is provided in the supplemental material. Tab.~\ref{tab:boundary_methods} summarizes existing methods addressing discontinuity. However, these methods are not universally continuous. To comprehensively resolve the problem of discontinuity, we propose COBB, which ensures both encoding continuity and decoding continuity.

\begin{figure*}[t!]
    \centering
    \subfloat[Base Parameters]{
        \begin{minipage}[b]{.19\textwidth}
            \centering
            \includegraphics[width=0.9\linewidth]{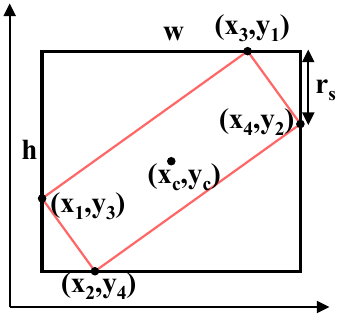}		
        \end{minipage}%
    }%
    \subfloat[Four types of OBBs with the same $(x_c, y_c, w, h, r_s)$]{
        \begin{minipage}[b]{.76\textwidth}
            \centering
            \includegraphics[width=0.9\linewidth]{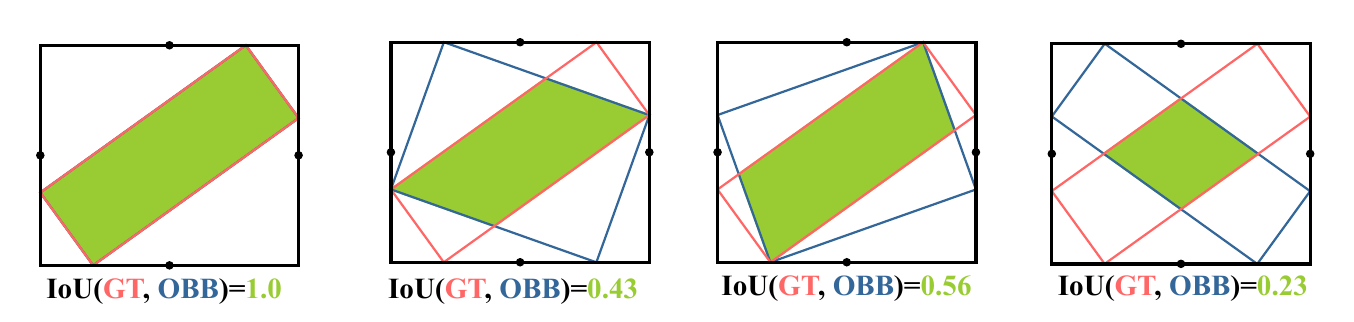}		
        \end{minipage}%
    }
    \vspace{-7pt}
    \caption{\textbf{Example of COBB}.
    COBB utilizes the outer HBB ($x_c, y_c, w, h$), sliding ratio $r_s$, and four IoU scores. (a) Example of the outer HBB and $r_s$. In this instance, $r_s=\frac{y_2-y_1}{h}$ when $w>h$, where $y_1$ and $y_2$ denote the two smaller y-coordinates among the four vertices of the OBB. (b) Using $x_c$, $y_c$, $w$, $h$, and $r_s$, along with the properties of similar triangles, we can derive and solve a system of equations to obtain the parameters for four OBBs (details provided in the supplemental material). Distinguishing between these OBBs is guided by the positional relationship between their vertices and the midpoints on each side.
    }
    \label{fig:four_type}
\vspace{-10pt}
\end{figure*}
\subsection{Our Continuous Representation for OBB}
\label{sec:method_details}
Note that the outer HBB and the area of an OBB undergo continuous changes during shape transformations. Consequently, we sought to represent an OBB with a 5-dimensional vector, $(x_c, y_c, w, h, r_a)$. Here, $(x_c, y_c)$, $w$, and $h$ refer to the center point, width, and height of the outer HBB, respectively, while $r_a$ is the acreage ratio of the OBB relative to its outer HBB.

It can be proven that only a pair of symmetrical OBBs shares the same $(x_c, y_c, w, h, r_a)$ (detailed proof is provided in the supplemental material). However, directly computing OBBs from $(x_c, y_c, w, h, r_a)$ is a complex process. To address this challenge, we introduce a sliding ratio, $r_s$, to estimate $r_a$, defined as follows.
\begin{equation}
\small
r_s=
\begin{cases}
\frac{x_2-x_1}{w} & w<h, \\
\frac{y_2-y_1}{h} & w\geq h,\\
\end{cases}
\label{eq:rs}
\end{equation}
where the x-coordinates of four vertices of the OBB are sorted as $x_1, x_2, x_3, x_4$ from small to large, and y-coordinates are sorted as $y_1, y_2, y_3, y_4$. It can be proved that $r_s$ can be computed as $r_s=f(\min(r_a, 1-r_a))$, where $f:[0, 0.5]\rightarrow [0, 0.5]$ is a continuous strictly increasing map (proof provided in the supplemental material). This implies the $r_s$ changes continuously as OBBs transform.

However, as shown in Fig.~\ref{fig:four_type}, a total of four different OBBs can be encoded into the same $(x_c, y_c, w, h, r_s)$, leading to potential DA. To mitigate DA, we utilize Intersection over Unions (IoUs) between the target OBB and the four OBBs as scores for classification. Importantly, these IoUs can be directly computed using $(x_c, y_c, w, h, r_s)$ and the classification of the target OBB, eliminating the need for complex computations involving IoU between arbitrary OBBs. The detailed computation process is provided in the supplemental material.

Finally, $(x_c, y_c, w, h, r_s, s_0, s_1, s_2, s_3)$ will be considered as a continuous representation of OBBs, where $s_0$, $s_1$, $s_2$, $s_3$ are IoUs between the target OBB and the four OBBs with the same $(x_c, y_c, w, h, r_s)$.

By reversing the above process, a 9-dimensional vector is decoded into a single OBB. Without loss of generality, assuming $w \geq h$, exploiting the properties of similar triangles allows the computation of $x_2-x_1$ and $y_2-y_1$:
\begin{equation}
\small
\begin{aligned}
y_2-y_1&= r_s h,\\
x_2-x_1&=\frac{1-\sqrt{1-4\cdot\frac{h^2}{w^2}\cdot r_s(1-r_s)}}{2}w.
\end{aligned}
\label{eq:rev_rs}
\end{equation}
The classification with the highest IoU score determines the style of the generated OBB. Using the HBB, $x_2-x_1$, $y_2-y_1$, and the style, the coordinates of the OBB's four vertices can be easily computed. The detailed computation process is provided in supplemental material.

\subsection{Implementing COBB in OOD Models}
\label{sec:method_target}
Most models use the bias between the ground truth and the assigned proposal as the regression target to take advantage of the priori information of proposals. In our method, for horizontal proposal region $(x_p, y_p, w_p, h_p)$, the regression target is computed as follows, same as Faster R-CNN~\cite{ren2015faster}:
\begin{equation}
\small
	\begin{aligned}
	t_x &= \frac{\Delta x}{w_p}, \
		t_y = \frac{\Delta y}{h_p}, \
		t_w = \ln\left(\frac{w}{w_p}\right), \
		t_h = \ln\left(\frac{h}{h_p}\right), \\
	\end{aligned}
\label{eq:smoothgv_bias}
\end{equation}
where $box_{t}=(t_x, t_y, t_w, t_h)$ is the target for the outer HBB, and $\Delta x = x_c-x_p$, $\Delta y = y_c-y_p$.

According to Eq.~\ref{eq:rs}, the value of $r_s$ lies within the range [0, 0.5]. To take advantage of this property, an effective way is to limit the range of prediction results, such as employing the sigmoid function. In this situation, the regression target for $r_s$ is computed as follows:
\begin{equation}
\small
    r_{sig}=2r_s.
    \label{eq:rsig}
\end{equation}
Another method is extending the domain of $r_s$ as follows:
\begin{equation}
\small
r_{ln}=
    \begin{cases}
        1+\log_2(r_s) & r_a < 0.5,\\
        1+\log_2(1-r_s) & r_a\geq 0.5.
    \end{cases}
\label{eq:rln}
\end{equation}
Compared with $r_{sig}$, $r_{ln}$ exhibits increased sensitivity to $r_s$ when $r_a$ is small, aiding detectors in precisely predicting inclined thin objects. Based on the definition of the regression target of $r_s$, our methods fall into two classifications: COBB-sig and COBB-ln.

The regression target of IoU scores is defined as:
\begin{equation}
\small
    s_{t}=(s_0^\lambda, s_1^\lambda, s_2^\lambda, s_3^\lambda),
    \label{eq:starget}
\end{equation}
where $\lambda$ is a predefined constant to amplify the gap between scores of the ground truth and other classifications.

For models employing oriented proposal regions, we rotate the proposal region and target OBB around the center of the proposal region until its rotation angle becomes zero. Subsequently, we calculate the regression target as that for horizontal proposal regions. By reversing this process, OBBs can be easily recovered from the regression target and the oriented proposal regions. The detailed computation method is provided in the supplemental material.

In our approach, the loss function is defined as:
\begin{equation}
\small
\begin{aligned}
    L=&w_1L_{cls}+w_2L_{box}(box_{p}, box_{t})\\
    &+w_3L_{r}(r_{p}, r_{t})+w_4L_{s}(s_{p}, s_{t}),
\end{aligned}
\label{eq:loss}
\end{equation}
where $box_p$, $r_p$, and $s_p$ denote predicted outer HBBs, predicted $r_s$, and predicted IoU scores, respectively. $r_t$ is the regression target of $r_s$, which is either $r_{sig}$ for COBB-sig or $r_{ln}$ for COBB-ln. $L_{cls}$ stands for the classification loss, which aligns with that of the baseline model (\eg cross-entropy loss for Faster R-CNN~\cite{ren2015faster}). $L_{box}$, $L_{r}$, and $L_{s}$ are Smooth L1 Loss~\cite{ren2015faster}. The hyperparameters $w_1$, $w_2$, $w_3$, and $w_4$ are predefined constants.

\subsection{Theoretical Guarantee of the Continuity}
\label{sec:continuity}
The COBB, as detailed in Sec.~\ref{sec:method_details}, theoretically ensures continuity under the metrics outlined in Sec.~\ref{sec:bdd_metric}. Here, we briefly elucidate the reasons behind the continuity, with detailed proofs provided in the supplemental material.

\textbf{Theoretical Analysis on Encoding Continuity:}
According to Sec.~\ref{sec:method_details}, $x_c$, $y_c$, $w$, $h$, and $r_s$ exhibit continuity concerning the outer HBB and the area of the target OBB. For unambiguous classifications, IoU scores remain continuous concerning $x_c$, $y_c$, $w$, $h$, and $r_s$. In cases of ambiguous OBB classifications, the IoU scores remain similar regardless of the classification. Consequently, the regression target produced by our method maintains continuity for both rotation and aspect ratio changes.

\textbf{Theoretical Analysis on Decoding Continuity:}
The OBB generation process in Sec.~\ref{sec:method_details} ensures precise reversal from the 9 parameters, mitigating inherent DI errors.

To avoid DA, the decoder must resist slight changes in its input. When IoU scores are fixed, the decoded four vertices remain continuous concerning $(x_c, y_c, w, h, r_s)$. Notably, when $w$ is similar to $h$, the changes in the ordering of values between $w$ and $h$ do not lead to DA, as $x_2-x_1$ is similar to $y_2 - y_1$. If $x_c$, $y_c$, $w$, $h$, and $r_s$ are fixed, and a slight perturbation in IoU scores results in a classification error, the IoU between the OBB before and after perturbation decoding is close to 1, adhering to the definition of IoU scores. In summary, our method exhibits resistance to perturbations in predicted results, thereby avoiding DA.

\subsection{Further Comparison with Peer Methods}
\textbf{Compared with Gliding Vertex:}
The Gliding Vertex method~\cite{xu2020gliding} represents an OBB by sliding the four vertices of its outer HBB. However, rotation continuity is compromised when the OBB is nearly horizontal. Moreover, its decoded results manifest as irregular quadrilaterals, and refining these into accurate OBBs introduces accuracy errors. In contrast, our methods ensure continuous prediction targets and loss values for nearly horizontal OBBs, and the decoded quadrilaterals consistently represent accurate OBBs.

\textbf{Compared with CSL-based methods:}
CSL-based methods~\cite{yang2020arbitrary, yang2021dense, wang2022gaussian, wang2022multigrained, zeng2023ars} discretize rotation angles, converting angle regression into an angle classification problem to address rotation discontinuity. However, angle discretization introduces DI problems and results in a heavy prediction layer. Additionally, most CSL-based methods do not maintain continuity in aspect ratio changes when dealing with square-like OBBs. In contrast, our method ensures encoding continuity in both rotation and aspect ratio changes without introducing DI. Furthermore, our approach encodes an OBB using only 9 parameters.

\begin{table}[t!]
	\renewcommand\arraystretch{1.4}
	\setlength\tabcolsep{4pt}
	\footnotesize
	\centering		
    \caption{\textbf{Open source OOD benchmarks.}}
    \label{tab:benchmarks}
	\vspace{-3mm}
    \resizebox{0.45\textwidth}{!}
{
		\begin{tabular}{cccccc}
\toprule
\textbf{Benchmark} & \textbf{ArialDet} & \textbf{OBBDet} & \textbf{AlphaRotate} \cite{yang2021alpharotate} & \textbf{MMRotate} \cite{zhou2022mmrotate} & \cellcolor{gray!20} \textbf{JDet} \\
\hline
DL library & PyTorch & PyTorch & Tensorflow & PyTorch & \cellcolor{gray!20} Jittor \\
Algorithm & 5 & 10 & 18 & 19 & \cellcolor{gray!20} 20\\
Dataset & 1 & 6 & 11 & 4 & \cellcolor{gray!20} 6 \\
\bottomrule

		\end{tabular}
 }
	\vspace{-3mm}
 \end{table}

\section{Benchmarking OOD under JDet}
\label{sec:JDet}
\subsection{Brief Description of JDet}
Our benchmark utilizes the \textbf{J}ittor object \textbf{DET}ection models library (JDet), an open-source library dedicated to object detection, particularly supporting OOD methods. Built on Jittor~\cite{hu2020jittor}, a deep learning framework, JDet facilitates the entire training and evaluation processes of object detection models. Preprocessing of diverse datasets precedes training or testing, ensuring a unified format. Various data augmentations, such as rotation and category balancing, are implemented as interchangeable and combinable modules. During testing, JDet supports diverse post-processing techniques for different datasets, with VOC2012~\cite{everingham2010pascal} serving as the implementation for evaluation. The library accommodates common object detection frameworks (e.g., Faster R-CNN~\cite{ren2015faster}) and operators for OOD (e.g., RRoI Align~\cite{ding2019learning}).

In total, JDet comprises 20 models and supports 6 datasets. A comparison between JDet and other open-source libraries is presented in Tab.~\ref{tab:benchmarks}.
\subsection{Components for Unified Benchmarking}
To mitigate variations between models, we categorized several modules and constructed OOD models by assembling these modules. The identified modules include:
\begin{itemize}
    \item \textbf{Backbone:} Extracts features from input images; most models employ ResNet~\cite{he2016deep} as the backbone network with FPN~\cite{lin2017feature} for feature extraction at different scales.
    \item \textbf{Anchor Generation:} Defines anchors for every pixel in the feature map.
    \item \textbf{Ground Truth Assignment:} Assigns ground truth bounding boxes to proposal regions based on their IoU.
    \item \textbf{Result Generation Network:} Neural networks for classifying anchors or proposal regions and regressing targets from regions. 
    \item \textbf{Encoder/Decoder:} Converts proposal regions into regression targets and model outputs into detection results.
    \item \textbf{Region of Interest Feature Extraction:} Extracts features of proposal regions for further detection and refinement, focusing on object-level details compared to image-level backbone features.
    \item \textbf{Loss Function:} Most models employ cross-entropy loss for classification and L1 loss for OBB prediction.
\end{itemize}
Implementing these modules consistently enhances the uniformity and comparability of the benchmarked models.

\begin{table}[t!]
	\renewcommand\arraystretch{1.4}
	\setlength\tabcolsep{4pt}
	\footnotesize
	\centering		
	\caption{\textbf{$\text{mAP}$ of models in JDet benchmark on DOTA-v1.0.}}
	\vspace{-3mm}
    \resizebox{0.48\textwidth}{!}
{
	\begin{tabular}{ccccc}
		\toprule
		\textbf{Model} & \textbf{Venue} & \textbf{$\text{mAP}_{50}$} & \textbf{$\text{mAP}_{75}$} & \textbf{$\text{mAP}_{50:95}$} \\ \hline
H2RBox~\cite{yang2023h2rbox} & ICLR'23 & 67.62 & 35.48 & 36.67 \\
CSL~\cite{yang2020arbitrary} & ECCV'20 & 67.99 & 34.51 & 36.43 \\
RSDet~\cite{qian2021learning} & AAAI'21 & 68.41 & 36.93 & 37.91\\
RetinaNet~\cite{lin2017focal} & ICCV'17 & 68.18 & 36.84 & 38.15\\
KLD~\cite{yang2021learning} & NeurIPS'21 & 68.75 & 38.68 & 39.29\\
KFIoU~\cite{yang2022kfiou} & ICLR'23 & 68.99 & 35.00 & 37.59\\
GWD~\cite{yang2021rethinking} & ICML'21 & 69.02 & 38.48 & 39.62\\
FCOS~\cite{tian2019fcos} & ICCV'19 & 70.37 & 39.78 & 40.25 \\
ATSS~\cite{zhang2020bridging} & CVPR'20 & 72.44 & 39.81 & 41.08 \\
$\text{S}^2\text{A-Net}$~\cite{han2021align} & TGRS'21 & 73.95 & 37.14 & 39.89\\
Faster R-CNN~\cite{ren2015faster} & NeurIPS'15 & 73.01 & 40.13 & 41.33\\
Gliding Vertex~\cite{xu2020gliding} & TPAMI'20 & 73.31 & 41.62 & 41.57\\
RoI Trans.~\cite{ding2019learning} & CVPR'19 & 75.59 & 48.54 & 46.35\\
Oriented R-CNN~\cite{xie2021oriented} & ICCV'21 & 75.11 & 47.48 & 45.20\\
ReDet~\cite{han2021redet} & CVPR'21 & 76.38 & 50.83 & 47.08\\
\rowcolor{gray!20} Ours (RoI Trans. based) & - & \textbf{76.53} & 50.41 & 46.97 \\
\rowcolor{gray!20} Ours (ReDet based) & - & 76.52 & \textbf{51.38} & \textbf{47.67}\\
		\bottomrule
	\end{tabular}
 }
	\vspace{-3mm}
	\label{tab:base}
 \end{table}

\subsection{Detection Models in the Benchmark}
Experiments are conducted on multiple models within our benchmark framework, all subject to uniform conditions. The baseline models chosen for comparison were Rotated Faster R-CNN~\cite{ren2015faster} and Rotated RetinaNet~\cite{lin2017focal}. To minimize implementation discrepancies, most of the other models are implemented with minimal alterations to their corresponding baseline architectures.

For fairness, we standardized data processing and training settings across different experiments, following the detailed settings outlined in Sec.~\ref{sec:datasets}. The experimental results on DOTA-v1.0 are presented in Tab.~\ref{tab:base}, with additional results available in the supplemental material.

\section{Experiments}
\label{sec:experiments}
\subsection{Datasets and Implementation Details}
\label{sec:datasets}

\textbf{DOTA}~\cite{xia2018dota} is a dataset for remote sensing object detection. We evaluated models on DOTA-v1.0 and DOTA-v1.5. DOTA-v1.0 comprises 2,806 aerial images whose resolution is between 800$\times$ 800 and 4,000$\times$ 4,000, and a total of 188,282 target instances are annotated, covering 15 common categories. DOTA-v1.5 maintains the same image and dataset segmentation as DOTA-v1.0 but introduces labeling for extremely small objects (less than 10 pixels) and incorporates the container crane (CC) category.

\begin{figure}[t!]
    \centering
    \subfloat[rotated Faster R-CNN + KLD]{
        \begin{minipage}[b]{.23\textwidth}
            \centering
            \includegraphics[width=0.95\linewidth]{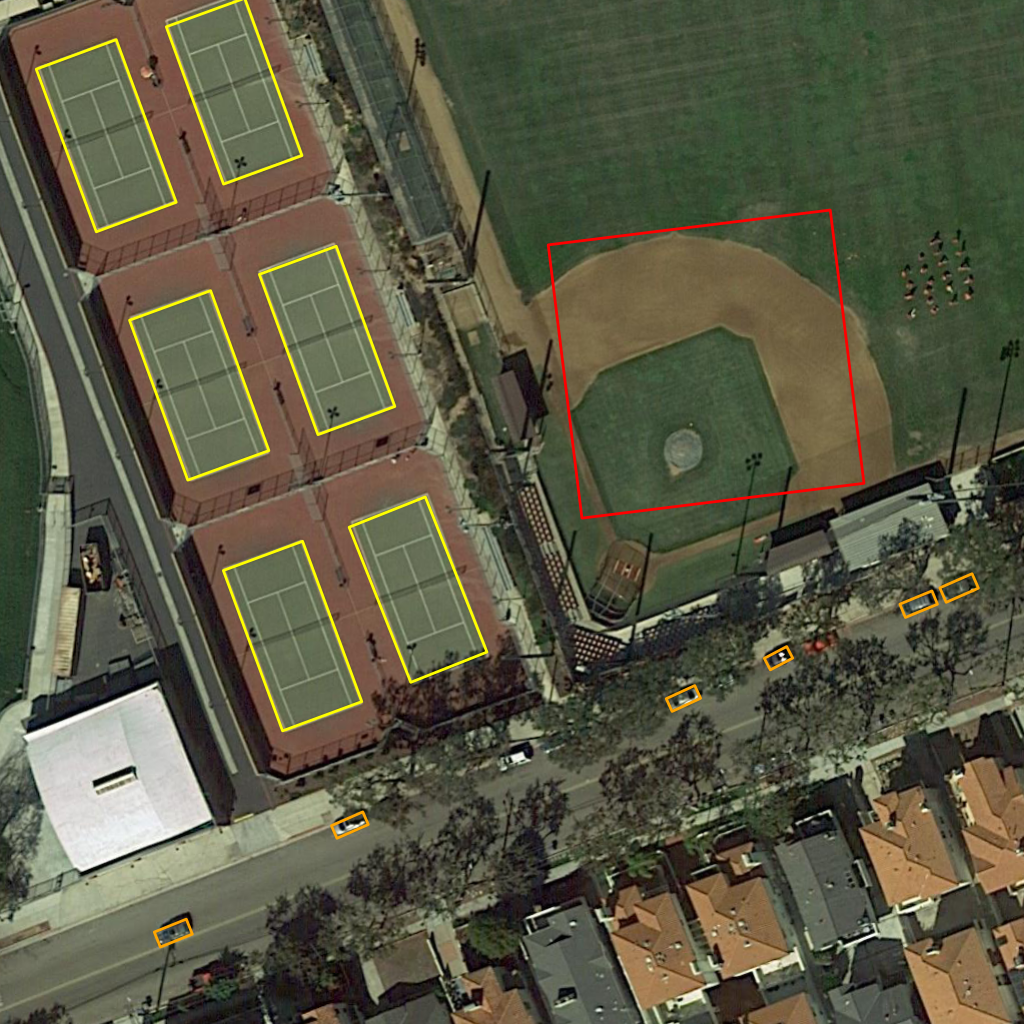}		
        \end{minipage}%
    }
    \subfloat[rotated Faster R-CNN + Ours]{
        \begin{minipage}[b]{.23\textwidth}
            \centering
            \includegraphics[width=0.95\linewidth]{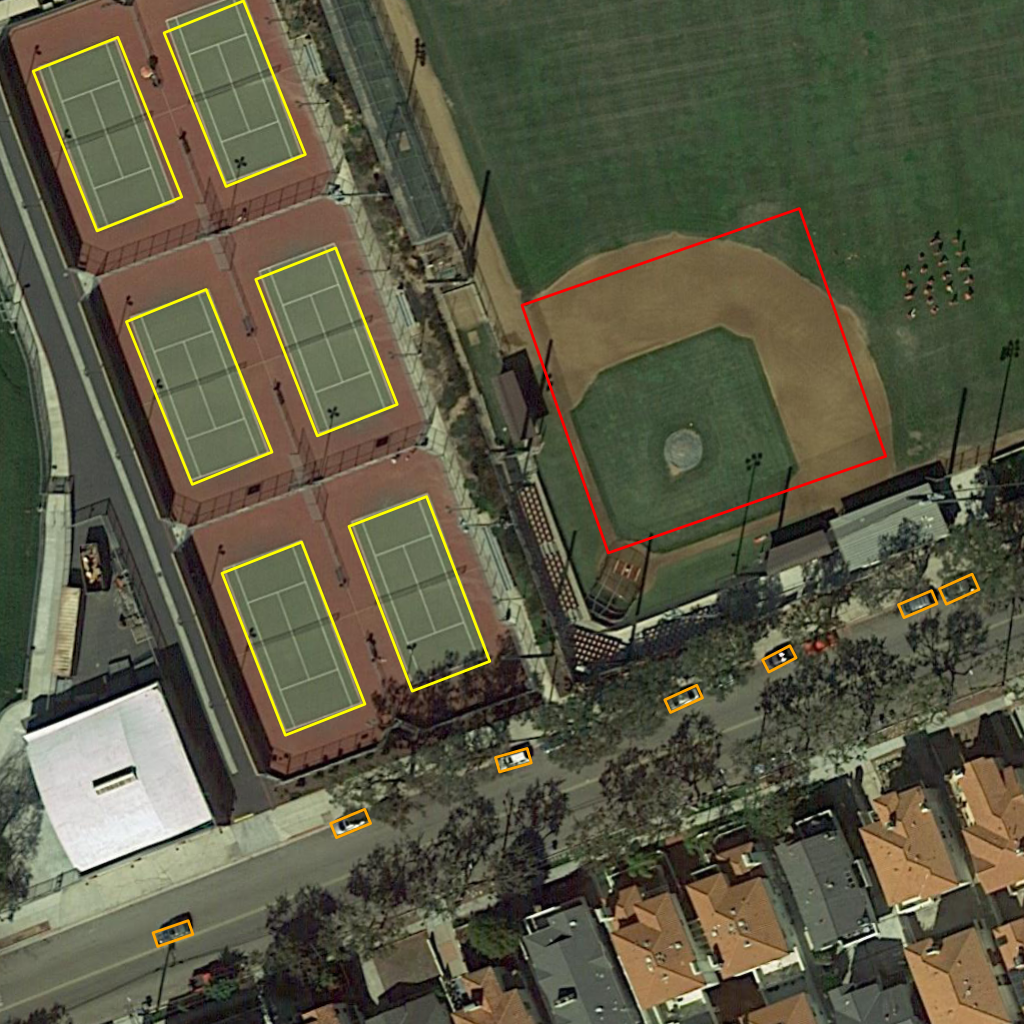}
        \end{minipage}%
    }
    \vspace{-5pt}
    \caption{\textbf{Visual results of KLD~\cite{yang2021learning} and ours.} Due to DA, KLD struggles to accurately predict the orientation of square-like objects. In contrast, our COBB circumvents DA, enhancing its precision in predicting the orientation of square-like objects. }
    \label{fig:visualization}
\end{figure}

\begin{table}[t!]
	\renewcommand\arraystretch{1.4}
	\setlength\tabcolsep{4pt}
	\footnotesize
	\centering		
    \caption{\textbf{Comparison: IoU Scores vs. One-hot Coding.}
The experimental setup remains consistent with Rotated Faster R-CNN + COBB-sig, except for the OBB classification scores. All experiments were performed on DOTA-v1.0.}
    \label{tab:one-hot}
	\vspace{-2mm}
    \resizebox{0.32\textwidth}{!}
{
		\begin{tabular}{cccccc}
\toprule
\textbf{Scores} &  \textbf{$\text{mAP}_{50}$} & \textbf{$\text{mAP}_{75}$} & \textbf{$\text{mAP}_{50:95}$} \\
\hline
One-hot & 73.46 & 43.76 & 42.90\\
IoU scores & 74.00 & 44.03 & 43.29\\
\bottomrule

		\end{tabular}
 }
	% \vspace{-5pt}
 \end{table}

 \begin{table}[t!]
	\renewcommand\arraystretch{1.4}
	\setlength\tabcolsep{4pt}
	\footnotesize
	\centering		
    \caption{\textbf{Comparison of COBB across Proposal Types.} Experiments were performed on DOTA-v1.0 using RoI Transformer, a model that incorporates both Horizontal Proposals (HPs) and Oriented Proposals (OPs), as the baseline.}
    \label{tab:roitrans_ablation}
	\vspace{-2mm}
    \resizebox{0.4\textwidth}{!}
{
		\begin{tabular}{ccccc}
\toprule
 \textbf{HPs} & \textbf{OPs} &  \textbf{$\text{mAP}_{50}$} & \textbf{$\text{mAP}_{75}$} & \textbf{$\text{mAP}_{50:95}$} \\
\hline
 - & - & 75.59 & 48.54 & 46.35\\
 COBB-ln & - & 76.27 & 50.23 & 47.06\\
 - & COBB-ln & 76.10 & 48.11 & 46.32\\
 \rowcolor{gray!20} COBB-ln & COBB-ln & 76.53 & 50.41 & 46.97\\
\bottomrule

		\end{tabular}
 } 
 \end{table}

\begin{table}[t!]
	\renewcommand\arraystretch{1.4}
	\setlength\tabcolsep{4pt}
	\footnotesize
	\centering		
    \caption{\textbf{Comparing Different Regression Targets.} The experimental setup mirrors that of Rotated Faster R-CNN + COBB-ln. One approach utilizes $r_a$, representing the OBB's acreage ratio concerning its outer HBB, for Regression Target (RT) calculation, while the alternative method uses $r_s$, the sliding ratio. All experiments were conducted on DOTA-v1.0. }
    \label{tab:r_target}
	\vspace{-2mm}
    \resizebox{0.32\textwidth}{!}
{
		\begin{tabular}{ccccc}
\toprule
 \textbf{RT} &  \textbf{$\text{mAP}_{50}$} & \textbf{$\text{mAP}_{75}$} & \textbf{$\text{mAP}_{50:95}$} \\
\hline
 Using $r_a$ & 74.13 & 43.31 & 42.94\\
 Using $r_s$ & 74.44 & 44.08 & 43.53\\
\bottomrule

		\end{tabular}
 }
	% \vspace{-8pt}
 \end{table}

\begin{table*}[t!]
	\renewcommand\arraystretch{1.4}
	\setlength\tabcolsep{4pt}
	\footnotesize
	\centering		
	\caption{\textbf{$\text{mAP}$ across datasets.} COBB-sig takes $r_{sig}$ for horizontal proposals, and COBB-ln takes $r_{ln}$ for horizontal ones. COBB-ln-sig takes $r_{ln}$ for horizontal ones, and $r_{sig}$ for rotated ones. The definition of COBB-ln-ln and COBB-sig-sig is similar.}
	\vspace{-3mm}
    \resizebox{\textwidth}{!}
{
	\begin{tabular}{cccccccccccccc}
		\toprule
		\multirow{2}*{Models} &  \multicolumn{3}{c}{DOTA-v1.0}  &  \multicolumn{3}{c}{DOTA-v1.5} & \multicolumn{2}{c}{DIOR} & \multicolumn{2}{c}{HRSC2016} & FAIR1M-1.0 & FAIR1M-2.0   \\   
		&  $\text{mAP}_{50}$ & $\text{mAP}_{75}$ & $\text{mAP}_{50:95}$&  $\text{mAP}_{50}$ & $\text{mAP}_{75}$ & $\text{mAP}_{50:95}$ &$\text{mAP}_{50}$ & $\text{mAP}_{75}$ & $\text{mAP}_{50}$ & $\text{mAP}_{75}$ & $\text{mAP}_{50}$ & $\text{mAP}_{50}$ \\ \hline
Rotated Faster R-CNN~\cite{ren2015faster} & 73.01 & 40.13 & 41.33 & 63.52 & 35.36 & 35.96 & 60.64 & 35.26 & 83.34 & 31.64 & 35.16 & 40.16\\
Gliding Vertex~\cite{xu2020gliding} & 73.31 & 41.62 & 41.57 & 63.12 & 36.98 & 36.32 & 61.49 & 36.24 & 92.23 & 58.52 & 36.37 & 40.82\\
\rowcolor{gray!20} +COBB-sig & 74.00 & 44.03 & 43.29 & 64.03 & 36.88 & 37.17 & 62.28 & \textbf{37.70} & 92.69 & 68.87 & \textbf{36.81} & 41.11\\
\rowcolor{gray!20} +COBB-ln & \textbf{74.44} & \textbf{44.08} & \textbf{43.53} & \textbf{64.35} & \textbf{37.62} & \textbf{37.30} & \textbf{62.58} & 37.55 & \textbf{92.71} & \textbf{72.29} & 36.53 & \textbf{41.23}\\\hline
RoI Trans.~\cite{ding2019learning} & 75.59 & 48.54 & 46.35 & 65.69 & 41.76 & 40.36 & 66.09 & 44.26 & 96.73 & 88.76 & 39.31 & 43.93\\
\rowcolor{gray!20} +COBB-sig-sig & 76.49 & 50.26 & 46.63 & 65.88 & 42.76 & 40.85 & 66.72 & 45.01 & 96.72 & 90.60 & 39.61 & 44.42\\
\rowcolor{gray!20} +COBB-ln-sig & \textbf{76.55} & 49.91 & 46.68 & \textbf{67.18} & 41.75 & 40.80 & 67.47 & \textbf{45.51} & 96.71 & 90.89 & \textbf{39.82} & \textbf{44.78}\\
\rowcolor{gray!20} +COBB-ln-ln & 76.53 & \textbf{50.41} & \textbf{46.97} & 66.66 & \textbf{43.29} & \textbf{40.96} & \textbf{67.53} & 45.27 & \textbf{97.19} & \textbf{91.35} & 39.66 & 44.54\\\hline
Oriented R-CNN~\cite{xie2021oriented} & 75.11 & 47.48 & 45.20 & 65.47 & 40.35 & 39.31 & 64.38 & 41.19 & 96.61 & 86.49 & 38.30 & 42.90\\
\rowcolor{gray!20} +COBB-sig & 75.52 & 48.35 & 45.61 & \textbf{66.25} & 41.34 & \textbf{40.04} & \textbf{65.65} & \textbf{42.78} & \textbf{96.77} & 87.43 & 38.81 & 43.31\\
\rowcolor{gray!20} +COBB-ln & \textbf{76.25} & \textbf{48.48} & \textbf{45.92} & 66.18 & \textbf{41.42} & 40.01 & 65.42 & 42.19 & 96.74 & \textbf{88.23} & \textbf{38.83} & \textbf{43.43}\\
		\bottomrule
	\end{tabular}

 }
	\vspace{-2mm}
	\label{tab:ablation_study}
 \end{table*}

\textbf{DIOR}~\cite{ding2021object} serves as a large-scale resource for remote sensing object detection, encompassing a total of 23,463 images, and spanning 20 distinct target categories. As stipulated in~\cite{ding2021object}, DIOR is partitioned into a training set of 11,725 images and a testing set of 11,738 images.

\textbf{HRSC2016}~\cite{liu2017high} is a ship detection dataset. Our training incorporates both the training and validation sets, while the test set is reserved for assessing model accuracy.

\textbf{FAIR1M}~\cite{sun2022fair1m}, designed for fine object detection in aerial images, consists of 5 categories with 37 subcategories. The dataset is available in two versions, -1.0 and -2.0. For FAIR1M-1.0, model training utilized the training set, and model evaluation was performed on the test set. For FAIR1M-2.0, models were trained on both the training and validation sets, with evaluation conducted on the test set.

All experiments were performed using a single NVIDIA RTX 3090. The models utilized ResNet-50~\cite{he2016deep} and FPN~\cite{lin2017feature} to extract multi-level feature maps. SGD optimization was employed during the training stage. Data augmentation included random flipping, with each image having a 50\% chance of horizontal flipping followed by a 50\% chance of vertical flipping.

\subsection{Ablation Study}
\textbf{Comparison between IoU Scores and One-hot Coding:}
In Sec.~\ref{sec:method_details}, we implemented IoU scores to differentiate OBBs sharing the same $(x_c, y_c, w, h, r_s)$. Alternatively, one-hot coding seems simpler for classification. We compared the two methods on rotated Faster R-CNN + COBB-sig, as recorded in Tab.~\ref{tab:one-hot}. The model's accuracy using one-hot coding is lower than that using IoU scores due to the discontinuity introduced by one-hot coding.

\textbf{Comparison of COBB across Proposal Types:}
The implementation of COBB on horizontal and oriented proposals is discussed in Sec.~\ref{sec:method_target}. To validate its effectiveness on both types, we conducted experiments on RoI Transformer~\cite{ding2019learning}, which employs both horizontal and oriented proposals. Results in Tab.~\ref{tab:roitrans_ablation} demonstrate integrating COBB enhances $\text{mAP}_{50}$ for both proposal types, with a more significant improvement observed in horizontal proposals. 

\textbf{Comparison between $r_a$ and $r_s$:}
In Sec.~\ref{sec:method_details}, we approximated $r_a$ with $r_s$. Further emphasizing the superiority of $r_s$ over $r_a$, experiments were conducted on Faster R-CNN, as shown in Tab.~\ref{tab:r_target}. The results illustrate that $r_s$ is more effective than $r_a$. This effectiveness stems from the complexity of recovering an OBB from the outer HBB and $r_a$, potentially leading to precision loss. Moreover, slight prediction errors on $r_a$ may cause significantly larger errors in the predicted OBB than errors caused by slight $r_s$ errors. Detailed insights are available in the supplemental material.

\subsection{Results and Analysis}
Detailed results on different datasets and detectors are presented in Tab.~\ref{tab:ablation_study}, with comprehensive ablation study details available in the supplemental material.

\textbf{Results on DOTA:}
The results on DOTA-v1.0  show that Gliding Vertex outperforms Faster R-CNN by 0.30\% in $\text{mAP}_{50}$ and 1.49\% in $\text{mAP}_{75}$. Despite its accuracy advantage, Gliding Vertex suffers from discontinuity and DI, limiting its overall accuracy. 

Our methods demonstrate superiority over existing approaches, especially in high-precision detection. Specifically, COBB outperforms Faster R-CNN by 1.21\%, 3.92\%, and 2.08\% in $\text{mAP}_{50}$, $\text{mAP}_{75}$, and $\text{mAP}_{50:95}$ on average. On RoI Transformer and Oriented R-CNN, our method also significantly enhances the accuracy. For RoI Transformer, it improves $\text{mAP}_{50}$, $\text{mAP}_{75}$, and $\text{mAP}_{50:95}$ by 0.93\%, 1.69\%, and 0.41\%, respectively. On Oriented R-CNN, it outperforms by 0.77\%, 0.94\%, and 0.56\%, respectively. Notably, our method exhibits clear advantages in $\text{mAP}_{75}$, signifying its capability in high-precision object detection, a result of continuous representation and avoidance of DI and DA.

On DOTA-v1.5, our method effectively boosts the baseline detector accuracy. On average, the performance gain is 1.89\%, 0.84\%, and 1.03\% in $\text{mAP}_{75}$ for Faster R-CNN, RoI Transformer, and Oriented R-CNN, respectively. This highlights its effectiveness in small object detection.

\textbf{Results on HRSC2016:}
HRSC2016 involves ship objects that are in large aspect ratios. As shown in Tab.~\ref{tab:ablation_study}, a substantial gap is observed between COBB-sig and COBB-ln. As discussed in Sec.~\ref{sec:method_target}, COBB-ln's advantage lies in its better capture of slight changes in $r_s$ for inclined large aspect ratio objects. Accordingly, COBB-ln outperforms COBB-sig by 3.42\% in $\text{mAP}_{75}$ and Gliding Vertex by 13.77\%. On RoI Transformer, COBB-ln-ln outperforms RoI Transformer by 2.59\%.

\textbf{Results on DIOR and FAIR1M:}
Experiments on less common datasets namely DIOR and FAIR1M, are also conducted. On DIOR, COBB-ln outperforms Faster R-CNN and RoI Transformer by 1.94\% and 1.44\%, respectively, in $\text{mAP}_{50}$, and by 2.29\% and 1.01\% in $\text{mAP}_{75}$. On FAIR1M-1.0 and FAIR1M-2.0, our method significantly improves baseline detectors.

\textbf{Visualization Results:}
Fig.~\ref{fig:visualization} visually compares the results of KLD~\cite{yang2021learning} and COBB. KLD's precision for the orientation of square-like objects is compromised by DA, whereas COBB accurately represents these objects, achieving strong performance by eliminating DA.

\section{Conclusion}
\label{sec:conclusion}
We have extensively shown the presence of boundary discontinuity in existing OOD models. 
To solve this problem, we have introduced COBB, an innovative continuous OBB representation method. Our experimental results showcase the effectiveness of our proposed method, achieving a notable improvement of 3.95\% in $\text{mAP}_{75}$ on Rotated Faster R-CNN applied to the DOTA Dataset, without employing any additional techniques. COBB also has limitations. The outer HBB, sliding ratio $r_s$, and IoU scores exhibit irregular variations during OBB rotation, restricting its impact on rotation-equivariant detectors (\eg ReDet~\cite{han2021redet}). Despite this, COBB proves effective in enhancing most OOD models by eliminating discontinuity.

\section{Ackownledgement}
The work was supported by the National Science and Technology Major Project under Grant 2021ZD0112902, the National Natural Science Foundation of China under Grant (62220106003, 62222607, 72342023), and the Research Grant of Beijing Higher Institution Engineering Research Center and Tsinghua-Tencent Joint Laboratory for Internet Innovation Technology.

{
    \small
    \bibliographystyle{ieeenat_fullname}
    \bibliography{arxiv}

\begin{thebibliography}{54}
\providecommand{\natexlab}[1]{#1}
\providecommand{\url}[1]{\texttt{#1}}
\expandafter\ifx\csname urlstyle\endcsname\relax
  \providecommand{\doi}[1]{doi: #1}\else
  \providecommand{\doi}{doi: \begingroup \urlstyle{rm}\Url}\fi

\bibitem[Chen et~al.(2020)Chen, Chen, Lin, See, Yu, Ke, and Yang]{chen2020piou}
Zhiming Chen, Kean Chen, Weiyao Lin, John See, Hui Yu, Yan Ke, and Cong Yang.
\newblock Piou loss: Towards accurate oriented object detection in complex environments.
\newblock In \emph{Computer Vision - {ECCV} 2020 - 16th European Conference, Glasgow, UK, August 23-28, 2020, Proceedings, Part {V}}, pages 195--211. Springer, 2020.

\bibitem[Ding et~al.(2019)Ding, Xue, Long, Xia, and Lu]{ding2019learning}
Jian Ding, Nan Xue, Yang Long, Gui{-}Song Xia, and Qikai Lu.
\newblock Learning roi transformer for oriented object detection in aerial images.
\newblock In \emph{{IEEE} Conference on Computer Vision and Pattern Recognition, {CVPR} 2019, Long Beach, CA, USA, June 16-20, 2019}, pages 2849--2858. Computer Vision Foundation / {IEEE}, 2019.

\bibitem[Ding et~al.(2022)Ding, Xue, Xia, Bai, Yang, Yang, Belongie, Luo, Datcu, Pelillo, and Zhang]{ding2021object}
Jian Ding, Nan Xue, Gui{-}Song Xia, Xiang Bai, Wen Yang, Michael~Ying Yang, Serge~J. Belongie, Jiebo Luo, Mihai Datcu, Marcello Pelillo, and Liangpei Zhang.
\newblock Object detection in aerial images: {A} large-scale benchmark and challenges.
\newblock \emph{{IEEE} Trans. Pattern Anal. Mach. Intell.}, 44\penalty0 (11):\penalty0 7778--7796, 2022.

\bibitem[Everingham et~al.(2010)Everingham, Gool, Williams, Winn, and Zisserman]{everingham2010pascal}
Mark Everingham, Luc~Van Gool, Christopher K.~I. Williams, John~M. Winn, and Andrew Zisserman.
\newblock The pascal visual object classes {(VOC)} challenge.
\newblock \emph{Int. J. Comput. Vis.}, 88\penalty0 (2):\penalty0 303--338, 2010.

\bibitem[Fu et~al.(2022)Fu, Jiang, Ji, Zhou, Zhao, and Fan]{fu2022light}
Keren Fu, Yao Jiang, Ge{-}Peng Ji, Tao Zhou, Qijun Zhao, and Deng{-}Ping Fan.
\newblock Light field salient object detection: {A} review and benchmark.
\newblock \emph{Computational Visual Media}, 8\penalty0 (4):\penalty0 509--534, 2022.

\bibitem[Guo et~al.(2022)Guo, Xu, Liu, Liu, Jiang, Mu, Zhang, Martin, Cheng, and Hu]{guo2022attention}
Meng{-}Hao Guo, Tian{-}Xing Xu, Jiang{-}Jiang Liu, Zheng{-}Ning Liu, Peng{-}Tao Jiang, Tai{-}Jiang Mu, Song{-}Hai Zhang, Ralph~R. Martin, Ming{-}Ming Cheng, and Shi{-}Min Hu.
\newblock Attention mechanisms in computer vision: {A} survey.
\newblock \emph{Computational Visual Media}, 8\penalty0 (3):\penalty0 331--368, 2022.

\bibitem[Han et~al.(2021)Han, Ding, Xue, and Xia]{han2021redet}
Jiaming Han, Jian Ding, Nan Xue, and Gui{-}Song Xia.
\newblock Redet: {A} rotation-equivariant detector for aerial object detection.
\newblock In \emph{{IEEE} Conference on Computer Vision and Pattern Recognition, {CVPR} 2021, virtual, June 19-25, 2021}, pages 2786--2795. Computer Vision Foundation / {IEEE}, 2021.

\bibitem[Han et~al.(2022)Han, Ding, Li, and Xia]{han2021align}
Jiaming Han, Jian Ding, Jie Li, and Gui{-}Song Xia.
\newblock Align deep features for oriented object detection.
\newblock \emph{{IEEE} Trans. Geosci. Remote. Sens.}, 60:\penalty0 1--11, 2022.

\bibitem[He et~al.(2016)He, Zhang, Ren, and Sun]{he2016deep}
Kaiming He, Xiangyu Zhang, Shaoqing Ren, and Jian Sun.
\newblock Deep residual learning for image recognition.
\newblock In \emph{2016 {IEEE} Conference on Computer Vision and Pattern Recognition, {CVPR} 2016, Las Vegas, NV, USA, June 27-30, 2016}, pages 770--778. {IEEE} Computer Society, 2016.

\bibitem[Hu et~al.(2020)Hu, Liang, Yang, Yang, and Zhou]{hu2020jittor}
Shi{-}Min Hu, Dun Liang, Guo{-}Ye Yang, Guo{-}Wei Yang, and Wen{-}Yang Zhou.
\newblock Jittor: a novel deep learning framework with meta-operators and unified graph execution.
\newblock \emph{Sci. China Inf. Sci.}, 63\penalty0 (12), 2020.

\bibitem[Lan et~al.(2022)Lan, Duan, Liu, Zhu, Xiong, Huang, and Xu]{lan2022arm3d}
Yuqing Lan, Yao Duan, Chenyi Liu, Chenyang Zhu, Yueshan Xiong, Hui Huang, and Kai Xu.
\newblock {ARM3D:} attention-based relation module for indoor 3d object detection.
\newblock \emph{Computational Visual Media}, 8\penalty0 (3):\penalty0 395--414, 2022.

\bibitem[Li et~al.(2023)Li, Hou, Zheng, Cheng, Yang, and Li]{li2023large}
Yuxuan Li, Qibin Hou, Zhaohui Zheng, Ming{-}Ming Cheng, Jian Yang, and Xiang Li.
\newblock Large selective kernel network for remote sensing object detection.
\newblock In \emph{{IEEE/CVF} International Conference on Computer Vision, {ICCV} 2023, Paris, France, October 1-6, 2023}, pages 16748--16759. {IEEE}, 2023.

\bibitem[Liao et~al.(2018)Liao, Zhu, Shi, Xia, and Bai]{Liao_2018_CVPR}
Minghui Liao, Zhen Zhu, Baoguang Shi, Gui{-}Song Xia, and Xiang Bai.
\newblock Rotation-sensitive regression for oriented scene text detection.
\newblock In \emph{2018 {IEEE} Conference on Computer Vision and Pattern Recognition, {CVPR} 2018, Salt Lake City, UT, USA, June 18-22, 2018}, pages 5909--5918. Computer Vision Foundation / {IEEE} Computer Society, 2018.

\bibitem[Lin et~al.(2017)Lin, Doll{\'{a}}r, Girshick, He, Hariharan, and Belongie]{lin2017feature}
Tsung{-}Yi Lin, Piotr Doll{\'{a}}r, Ross~B. Girshick, Kaiming He, Bharath Hariharan, and Serge~J. Belongie.
\newblock Feature pyramid networks for object detection.
\newblock In \emph{2017 {IEEE} Conference on Computer Vision and Pattern Recognition, {CVPR} 2017, Honolulu, HI, USA, July 21-26, 2017}, pages 936--944. {IEEE} Computer Society, 2017.

\bibitem[Lin et~al.(2020)Lin, Goyal, Girshick, He, and Doll{\'{a}}r]{lin2017focal}
Tsung{-}Yi Lin, Priya Goyal, Ross~B. Girshick, Kaiming He, and Piotr Doll{\'{a}}r.
\newblock Focal loss for dense object detection.
\newblock \emph{{IEEE} Trans. Pattern Anal. Mach. Intell.}, 42\penalty0 (2):\penalty0 318--327, 2020.

\bibitem[Liu et~al.(2016)Liu, Anguelov, Erhan, Szegedy, Reed, Fu, and Berg]{liu2016ssd}
Wei Liu, Dragomir Anguelov, Dumitru Erhan, Christian Szegedy, Scott~E. Reed, Cheng{-}Yang Fu, and Alexander~C. Berg.
\newblock {SSD:} single shot multibox detector.
\newblock In \emph{Computer Vision - {ECCV} 2016 - 14th European Conference, Amsterdam, The Netherlands, October 11-14, 2016, Proceedings, Part {I}}, pages 21--37. Springer, 2016.

\bibitem[Liu et~al.(2017{\natexlab{a}})Liu, Hu, Weng, and Yang]{liu2017rotated}
Zikun Liu, Jingao Hu, Lubin Weng, and Yiping Yang.
\newblock Rotated region based {CNN} for ship detection.
\newblock In \emph{2017 {IEEE} International Conference on Image Processing, {ICIP} 2017, Beijing, China, September 17-20, 2017}, pages 900--904. {IEEE}, 2017{\natexlab{a}}.

\bibitem[Liu et~al.(2017{\natexlab{b}})Liu, Yuan, Weng, and Yang]{liu2017high}
Zikun Liu, Liu Yuan, Lubin Weng, and Yiping Yang.
\newblock A high resolution optical satellite image dataset for ship recognition and some new baselines.
\newblock In \emph{Proceedings of the 6th International Conference on Pattern Recognition Applications and Methods, {ICPRAM} 2017, Porto, Portugal, February 24-26, 2017}, pages 324--331. SciTePress, 2017{\natexlab{b}}.

\bibitem[Ma et~al.(2018{\natexlab{a}})Ma, Shao, Ye, Wang, Wang, Zheng, and Xue]{8323240}
Jianqi Ma, Weiyuan Shao, Hao Ye, Li Wang, Hong Wang, Yingbin Zheng, and Xiangyang Xue.
\newblock Arbitrary-oriented scene text detection via rotation proposals.
\newblock \emph{{IEEE} Trans. Multim.}, 20\penalty0 (11):\penalty0 3111--3122, 2018{\natexlab{a}}.

\bibitem[Ma et~al.(2018{\natexlab{b}})Ma, Shao, Ye, Wang, Wang, Zheng, and Xue]{ma2018arbitrary}
Jianqi Ma, Weiyuan Shao, Hao Ye, Li Wang, Hong Wang, Yingbin Zheng, and Xiangyang Xue.
\newblock Arbitrary-oriented scene text detection via rotation proposals.
\newblock \emph{{IEEE} Trans. Multim.}, 20\penalty0 (11):\penalty0 3111--3122, 2018{\natexlab{b}}.

\bibitem[Ming et~al.(2022)Ming, Miao, Zhou, Yang, and Dong]{ming2021optimization}
Qi Ming, Lingjuan Miao, Zhiqiang Zhou, Xue Yang, and Yunpeng Dong.
\newblock Optimization for arbitrary-oriented object detection via representation invariance loss.
\newblock \emph{{IEEE} Geosci. Remote. Sens. Lett.}, 19:\penalty0 1--5, 2022.

\bibitem[Nie and Huang(2023)]{nie2022multi}
Guangtao Nie and Hua Huang.
\newblock Multi-oriented object detection in aerial images with double horizontal rectangles.
\newblock \emph{{IEEE} Trans. Pattern Anal. Mach. Intell.}, 45\penalty0 (4):\penalty0 4932--4944, 2023.

\bibitem[Qian et~al.(2021)Qian, Yang, Peng, Yan, and Guo]{qian2021learning}
Wen Qian, Xue Yang, Silong Peng, Junchi Yan, and Yue Guo.
\newblock Learning modulated loss for rotated object detection.
\newblock In \emph{Thirty-Fifth {AAAI} Conference on Artificial Intelligence, {AAAI} 2021, Thirty-Third Conference on Innovative Applications of Artificial Intelligence, {IAAI} 2021, The Eleventh Symposium on Educational Advances in Artificial Intelligence, {EAAI} 2021, Virtual Event, February 2-9, 2021}, pages 2458--2466. {AAAI} Press, 2021.

\bibitem[Redmon et~al.(2016)Redmon, Divvala, Girshick, and Farhadi]{redmon2016you}
Joseph Redmon, Santosh~Kumar Divvala, Ross~B. Girshick, and Ali Farhadi.
\newblock You only look once: Unified, real-time object detection.
\newblock In \emph{2016 {IEEE} Conference on Computer Vision and Pattern Recognition, {CVPR} 2016, Las Vegas, NV, USA, June 27-30, 2016}, pages 779--788. {IEEE} Computer Society, 2016.

\bibitem[Ren et~al.(2015)Ren, He, Girshick, and Sun]{ren2015faster}
Shaoqing Ren, Kaiming He, Ross~B. Girshick, and Jian Sun.
\newblock Faster {R-CNN:} towards real-time object detection with region proposal networks.
\newblock In \emph{Advances in Neural Information Processing Systems 28: Annual Conference on Neural Information Processing Systems 2015, December 7-12, 2015, Montreal, Quebec, Canada}, pages 91--99, 2015.

\bibitem[Sun et~al.(2022)Sun, Wang, Yan, Xu, Wang, Diao, Chen, Li, Feng, Xu, et~al.]{sun2022fair1m}
Xian Sun, Peijin Wang, Zhiyuan Yan, Feng Xu, Ruiping Wang, Wenhui Diao, Jin Chen, Jihao Li, Yingchao Feng, Tao Xu, et~al.
\newblock Fair1m: A benchmark dataset for fine-grained object recognition in high-resolution remote sensing imagery.
\newblock \emph{ISPRS Journal of Photogrammetry and Remote Sensing}, 184:\penalty0 116--130, 2022.

\bibitem[Tian et~al.(2019)Tian, Shen, Chen, and He]{tian2019fcos}
Zhi Tian, Chunhua Shen, Hao Chen, and Tong He.
\newblock {FCOS:} fully convolutional one-stage object detection.
\newblock In \emph{2019 {IEEE/CVF} International Conference on Computer Vision, {ICCV} 2019, Seoul, Korea (South), October 27 - November 2, 2019}, pages 9626--9635. {IEEE}, 2019.

\bibitem[Wang et~al.(2022{\natexlab{a}})Wang, Huang, Chen, Song, and Li]{wang2022multigrained}
Hao Wang, Zhanchao Huang, Zhengchao Chen, Ying Song, and Wei Li.
\newblock Multigrained angle representation for remote-sensing object detection.
\newblock \emph{{IEEE} Trans. Geosci. Remote. Sens.}, 60:\penalty0 1--13, 2022{\natexlab{a}}.

\bibitem[Wang et~al.(2022{\natexlab{b}})Wang, Li, and Bi]{wang2022gaussian}
Jian Wang, Fan Li, and Haixia Bi.
\newblock Gaussian focal loss: Learning distribution polarized angle prediction for rotated object detection in aerial images.
\newblock \emph{{IEEE} Trans. Geosci. Remote. Sens.}, 60:\penalty0 1--13, 2022{\natexlab{b}}.

\bibitem[Wang et~al.(2022{\natexlab{c}})Wang, Xie, Li, Fan, Song, Liang, Lu, Luo, and Shao]{wang2022pvt}
Wenhai Wang, Enze Xie, Xiang Li, Deng{-}Ping Fan, Kaitao Song, Ding Liang, Tong Lu, Ping Luo, and Ling Shao.
\newblock {PVT} v2: Improved baselines with pyramid vision transformer.
\newblock \emph{Computational Visual Media}, 8\penalty0 (3):\penalty0 415--424, 2022{\natexlab{c}}.

\bibitem[Wei et~al.(2020)Wei, Zhang, Chang, Li, Wang, and Sun]{wei2020oriented}
Haoran Wei, Yue Zhang, Zhonghan Chang, Hao Li, Hongqi Wang, and Xian Sun.
\newblock Oriented objects as pairs of middle lines.
\newblock \emph{ISPRS Journal of Photogrammetry and Remote Sensing}, 169:\penalty0 268--279, 2020.

\bibitem[Xia et~al.(2018)Xia, Bai, Ding, Zhu, Belongie, Luo, Datcu, Pelillo, and Zhang]{xia2018dota}
Gui{-}Song Xia, Xiang Bai, Jian Ding, Zhen Zhu, Serge~J. Belongie, Jiebo Luo, Mihai Datcu, Marcello Pelillo, and Liangpei Zhang.
\newblock {DOTA:} {A} large-scale dataset for object detection in aerial images.
\newblock In \emph{2018 {IEEE} Conference on Computer Vision and Pattern Recognition, {CVPR} 2018, Salt Lake City, UT, USA, June 18-22, 2018}, pages 3974--3983. Computer Vision Foundation / {IEEE} Computer Society, 2018.

\bibitem[Xie et~al.(2021)Xie, Cheng, Wang, Yao, and Han]{xie2021oriented}
Xingxing Xie, Gong Cheng, Jiabao Wang, Xiwen Yao, and Junwei Han.
\newblock Oriented {R-CNN} for object detection.
\newblock In \emph{2021 {IEEE/CVF} International Conference on Computer Vision, {ICCV} 2021, Montreal, QC, Canada, October 10-17, 2021}, pages 3500--3509. {IEEE}, 2021.

\bibitem[Xu et~al.(2023)Xu, Liu, Xu, Ma, Zhu, Yan, and Dai]{xu2023rethinking}
Hang Xu, Xinyuan Liu, Haonan Xu, Yike Ma, Zunjie Zhu, Chenggang Yan, and Feng Dai.
\newblock Rethinking boundary discontinuity problem for oriented object detection.
\newblock \emph{CoRR}, abs/2305.10061, 2023.

\bibitem[Xu et~al.(2021)Xu, Fu, Wang, Wang, Chen, Xia, and Bai]{xu2020gliding}
Yongchao Xu, Mingtao Fu, Qimeng Wang, Yukang Wang, Kai Chen, Gui{-}Song Xia, and Xiang Bai.
\newblock Gliding vertex on the horizontal bounding box for multi-oriented object detection.
\newblock \emph{{IEEE} Trans. Pattern Anal. Mach. Intell.}, 43\penalty0 (4):\penalty0 1452--1459, 2021.

\bibitem[Yang et~al.(2023{\natexlab{a}})Yang, Li, Xiao, Mu, Martin, and Hu]{yang2023sampling}
Guo{-}Ye Yang, Xiang{-}Li Li, Zi{-}Kai Xiao, Tai{-}Jiang Mu, Ralph~R. Martin, and Shi{-}Min Hu.
\newblock Sampling equivariant self-attention networks for object detection in aerial images.
\newblock \emph{{IEEE} Trans. Image Process.}, 32:\penalty0 6413--6425, 2023{\natexlab{a}}.

\bibitem[Yang and Yan(2020)]{yang2020arbitrary}
Xue Yang and Junchi Yan.
\newblock Arbitrary-oriented object detection with circular smooth label.
\newblock In \emph{Computer Vision - {ECCV} 2020 - 16th European Conference, Glasgow, UK, August 23-28, 2020, Proceedings, Part {VIII}}, pages 677--694. Springer, 2020.

\bibitem[Yang et~al.(2019)Yang, Yang, Yan, Zhang, Zhang, Guo, Sun, and Fu]{yang2019scrdet}
Xue Yang, Jirui Yang, Junchi Yan, Yue Zhang, Tengfei Zhang, Zhi Guo, Xian Sun, and Kun Fu.
\newblock Scrdet: Towards more robust detection for small, cluttered and rotated objects.
\newblock In \emph{2019 {IEEE/CVF} International Conference on Computer Vision, {ICCV} 2019, Seoul, Korea (South), October 27 - November 2, 2019}, pages 8231--8240. {IEEE}, 2019.

\bibitem[Yang et~al.(2021{\natexlab{a}})Yang, Hou, Zhou, Wang, and Yan]{yang2021dense}
Xue Yang, Liping Hou, Yue Zhou, Wentao Wang, and Junchi Yan.
\newblock Dense label encoding for boundary discontinuity free rotation detection.
\newblock In \emph{{IEEE} Conference on Computer Vision and Pattern Recognition, {CVPR} 2021, virtual, June 19-25, 2021}, pages 15819--15829. Computer Vision Foundation / {IEEE}, 2021{\natexlab{a}}.

\bibitem[Yang et~al.(2021{\natexlab{b}})Yang, Yan, Ming, Wang, Zhang, and Tian]{yang2021rethinking}
Xue Yang, Junchi Yan, Qi Ming, Wentao Wang, Xiaopeng Zhang, and Qi Tian.
\newblock Rethinking rotated object detection with gaussian wasserstein distance loss.
\newblock In \emph{Proceedings of the 38th International Conference on Machine Learning, {ICML} 2021, 18-24 July 2021, Virtual Event}, pages 11830--11841. {PMLR}, 2021{\natexlab{b}}.

\bibitem[Yang et~al.(2021{\natexlab{c}})Yang, Yang, Yang, Ming, Wang, Tian, and Yan]{yang2021learning}
Xue Yang, Xiaojiang Yang, Jirui Yang, Qi Ming, Wentao Wang, Qi Tian, and Junchi Yan.
\newblock Learning high-precision bounding box for rotated object detection via kullback-leibler divergence.
\newblock In \emph{Advances in Neural Information Processing Systems 34: Annual Conference on Neural Information Processing Systems 2021, NeurIPS 2021, December 6-14, 2021, virtual}, pages 18381--18394, 2021{\natexlab{c}}.

\bibitem[Yang et~al.(2021{\natexlab{d}})Yang, Zhou, and Yan]{yang2021alpharotate}
Xue Yang, Yue Zhou, and Junchi Yan.
\newblock Alpharotate: {A} rotation detection benchmark using tensorflow.
\newblock \emph{CoRR}, abs/2111.06677, 2021{\natexlab{d}}.

\bibitem[Yang et~al.(2023{\natexlab{b}})Yang, Zhang, Li, Zhou, Wang, and Yan]{yang2023h2rbox}
Xue Yang, Gefan Zhang, Wentong Li, Yue Zhou, Xuehui Wang, and Junchi Yan.
\newblock H2rbox: Horizontal box annotation is all you need for oriented object detection.
\newblock In \emph{The Eleventh International Conference on Learning Representations, {ICLR} 2023, Kigali, Rwanda, May 1-5, 2023}. OpenReview.net, 2023{\natexlab{b}}.

\bibitem[Yang et~al.(2023{\natexlab{c}})Yang, Zhou, Zhang, Yang, Wang, Yan, Zhang, and Tian]{yang2022kfiou}
Xue Yang, Yue Zhou, Gefan Zhang, Jirui Yang, Wentao Wang, Junchi Yan, Xiaopeng Zhang, and Qi Tian.
\newblock The kfiou loss for rotated object detection.
\newblock In \emph{The Eleventh International Conference on Learning Representations, {ICLR} 2023, Kigali, Rwanda, May 1-5, 2023}. OpenReview.net, 2023{\natexlab{c}}.

\bibitem[Yi et~al.(2021)Yi, Wu, Liu, Huang, Qu, and Metaxas]{yi2021oriented}
Jingru Yi, Pengxiang Wu, Bo Liu, Qiaoying Huang, Hui Qu, and Dimitris~N. Metaxas.
\newblock Oriented object detection in aerial images with box boundary-aware vectors.
\newblock In \emph{{IEEE} Winter Conference on Applications of Computer Vision, {WACV} 2021, Waikoloa, HI, USA, January 3-8, 2021}, pages 2149--2158. {IEEE}, 2021.

\bibitem[Yu and Da(2023)]{yu2023phase}
Yi Yu and Feipeng Da.
\newblock Phase-shifting coder: Predicting accurate orientation in oriented object detection.
\newblock In \emph{{IEEE/CVF} Conference on Computer Vision and Pattern Recognition, {CVPR} 2023, Vancouver, BC, Canada, June 17-24, 2023}, pages 13354--13363. {IEEE}, 2023.

\bibitem[Yuan et~al.(2022)Yuan, Li, and Ma]{yuan2022feature}
Yuan Yuan, Zhiguo Li, and Dandan Ma.
\newblock Feature-aligned single-stage rotation object detection with continuous boundary.
\newblock \emph{{IEEE} Trans. Geosci. Remote. Sens.}, 60:\penalty0 1--11, 2022.

\bibitem[Zeng et~al.(2023)Zeng, Yang, Li, Chen, and Yan]{zeng2023ars}
Ying Zeng, Xue Yang, Qingyun Li, Yushi Chen, and Junchi Yan.
\newblock {ARS-DETR:} aspect ratio sensitive oriented object detection with transformer.
\newblock \emph{CoRR}, abs/2303.04989, 2023.

\bibitem[Zhang et~al.(2023)Zhang, Wang, Shen, Zhao, Zeng, Li, and Li]{zhang2023trigonometric}
Rufei Zhang, Yuqing Wang, Sheng Shen, Wei Zhao, Zhiliang Zeng, Nannan Li, and Dongjin Li.
\newblock Trigonometric-coded refined detector for high precision oriented object detection.
\newblock \emph{{IEEE} Geosci. Remote. Sens. Lett.}, 20:\penalty0 1--5, 2023.

\bibitem[Zhang et~al.(2020)Zhang, Chi, Yao, Lei, and Li]{zhang2020bridging}
Shifeng Zhang, Cheng Chi, Yongqiang Yao, Zhen Lei, and Stan~Z. Li.
\newblock Bridging the gap between anchor-based and anchor-free detection via adaptive training sample selection.
\newblock In \emph{2020 {IEEE/CVF} Conference on Computer Vision and Pattern Recognition, {CVPR} 2020, Seattle, WA, USA, June 13-19, 2020}, pages 9756--9765. Computer Vision Foundation / {IEEE}, 2020.

\bibitem[Zhao et~al.(2021)Zhao, Qu, Bu, Tan, and Guan]{zhao2021polardet}
Pengbo Zhao, Zhenshen Qu, Yingjia Bu, Wenming Tan, and Qiuyu Guan.
\newblock Polardet: A fast, more precise detector for rotated target in aerial images.
\newblock \emph{International Journal of Remote Sensing}, 42\penalty0 (15):\penalty0 5831--5861, 2021.

\bibitem[Zhao et~al.(2019)Zhao, Zheng, Xu, and Wu]{8627998}
Zhong{-}Qiu Zhao, Peng Zheng, Shou{-}tao Xu, and Xindong Wu.
\newblock Object detection with deep learning: {A} review.
\newblock \emph{{IEEE} Trans. Neural Networks Learn. Syst.}, 30\penalty0 (11):\penalty0 3212--3232, 2019.

\bibitem[Zhou et~al.(2021)Zhou, Fan, Cheng, Shen, and Shao]{zhou2021rgb}
Tao Zhou, Deng{-}Ping Fan, Ming{-}Ming Cheng, Jianbing Shen, and Ling Shao.
\newblock {RGB-D} salient object detection: {A} survey.
\newblock \emph{Computational Visual Media}, 7\penalty0 (1):\penalty0 37--69, 2021.

\bibitem[Zhou et~al.(2022)Zhou, Yang, Zhang, Wang, Liu, Hou, Jiang, Liu, Yan, Lyu, Zhang, and Chen]{zhou2022mmrotate}
Yue Zhou, Xue Yang, Gefan Zhang, Jiabao Wang, Yanyi Liu, Liping Hou, Xue Jiang, Xingzhao Liu, Junchi Yan, Chengqi Lyu, Wenwei Zhang, and Kai Chen.
\newblock Mmrotate: {A} rotated object detection benchmark using pytorch.
\newblock In \emph{{MM} '22: The 30th {ACM} International Conference on Multimedia, Lisboa, Portugal, October 10 - 14, 2022}, pages 7331--7334. {ACM}, 2022.

\end{thebibliography}
}

\clearpage
\setcounter{page}{1}
\maketitlesupplementary
\section{Futher Explain of Metrics for Continuity}\label{sec:futher_explain}
\label{sup: metrics}
Within Sec.~\ref{sec:bdd_metric}, we introduced diverse metrics for evaluating methodological continuity. These metrics encompass Target Rotation Continuity, Target Aspect Ratio Continuity, Loss Rotation Continuity, Loss Aspect Ratio Continuity, Decoding Completeness, and Decoding Robustness. While formal definitions were provided in Sec.~\ref{sec:bdd_metric}, this section delves deeper into their conceptual underpinnings.

\textbf{Target Rotation Continuity:}
This metric assesses whether each OBB is encoded into a sole prediction target, and if slight rotations induce gradual changes in the prediction target. Notably, PSC~\cite{yu2023phase} demonstrates target rotation continuity by utilizing phase-shifting coding, ensuring continuous encoding despite OBB orientation changes. Conversely, Gliding Vertex~\cite{xu2020gliding} exhibits notable deviations in target rotation continuity when minor rotations affect nearly horizontal OBBs, leading to abrupt changes in the prediction target.

\textbf{Target Aspect ratio Continuity:}
Here, the focus lies on determining if every OBB is encoded into a single prediction target and whether slight aspect ratio adjustments cause sudden changes in the prediction target. For instance, Gliding Vertex~\cite{xu2020gliding} maintains target aspect ratio continuity. However, PSC~\cite{yu2023phase} struggles to sustain aspect ratio continuity, particularly when dealing with square-shaped OBBs.

\textbf{Loss Continuity:}
Loss Continuity encompasses two distinct components: Loss Rotation Continuity and Loss Aspect Ratio Continuity. This metric evaluates whether minor rotation or aspect ratio changes result in abrupt fluctuations in the loss value. While some methods might falter in maintaining target rotation or aspect ratio continuity, they compensate by refining the loss function to ensure loss continuity. Notably, employing an L1 Loss function aids in preserving loss continuity for methods demonstrating target continuity.

\textbf{Decoding Completeness:}
This criterion mandates precise representation of all OBBs. Methods rooted in CSL paradigms, such as those discussed in \cite{yang2020arbitrary, yang2021dense, wang2022gaussian}, often fall short in achieving Decoding Completeness due to discretized angle predictions, leading to imprecise orientation estimations of OBBs within finite angle classifications. Notably, we consider methods based on Gaussian distribution (such as GWD~\cite{yang2021rethinking}, KLD~\cite{yang2021learning}, and KFIoU~\cite{yang2022kfiou}) satisfying Decoding Completeness because squares in various orientations can be possibly precisely decoded theoretically. However, in actual implementation, squares in only one orientation can be precisely decoded.

\textbf{Decoding Robustness:}
Decoding Robustness demands that decoded OBBs remain resilient to slight errors in their representation. An example illustrating dissatisfaction with decoding robustness is GWD~\cite{yang2021rethinking}, which assigns square-like OBBs in different orientations to the same Gaussian distribution, leading to imprecise predictions for square-like targets. 
Especially, DHRec~\cite{nie2022multi} encodes two symmetrical tilted slender OBBs into comparable representations, particularly when the aspect ratio is significantly large. Although this is uncommon, it suggests a propensity for the algorithm to confuse slender OBBs that are oriented in differing directions. Consequently, this observation leads us to conclude that DHRec does not fulfill the criteria for Decoding Robustness.

Target Rotation Continuity, Target Aspect ratio Continuity, Loss Rotation Continuity, and Loss Aspect Ratio Continuity are collectively referred to as Encoding Continuity, while Decoding Completeness and Decoding Robustness are collectively referred to as Decoding Continuity.

\section{Details of COBB}
\subsection{Derivation of Four OBBs with Identical Outer HBB and $r_s$}
\label{sup: four_type}

\begin{figure}[t]
\centering
\includegraphics[width=0.5\linewidth]{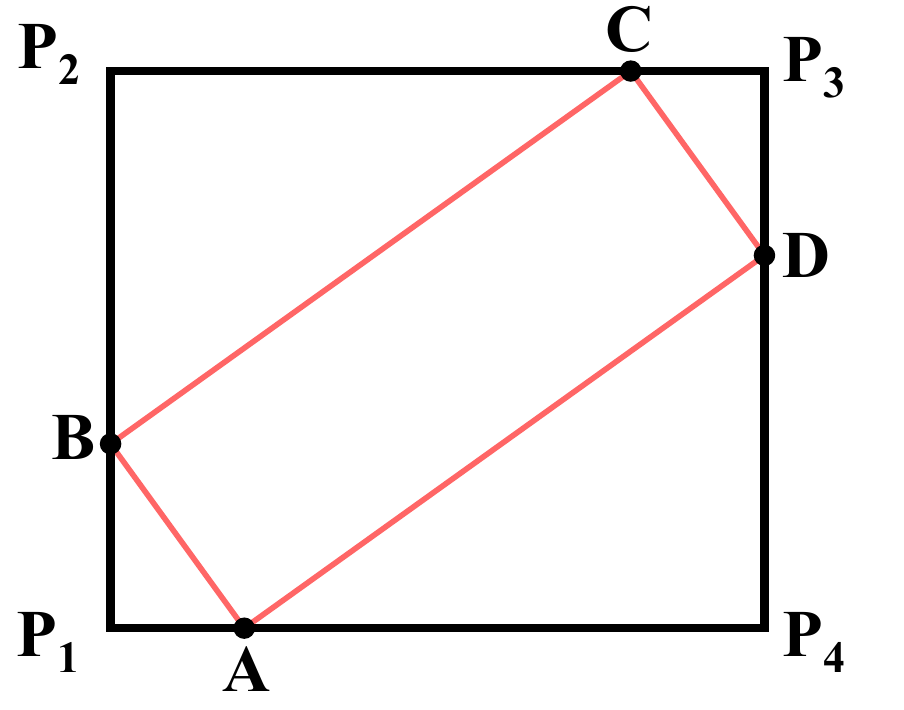}
\caption{\textbf{An OBB with Its outer HBB.}}
\label{fig: example_obb}
\end{figure}

We establish the outer HBB ($x_c, y_c, w, h$) of an OBB along with the sliding ratio $r_s$ in Sec.~\ref{sec:method_details}. Herein, we elaborate on the existence of precisely four OBBs sharing the same outer HBB and $r_s$.

Given a generic OBB characterized by the outer HBB $x_c, y_c, w, h$, and sliding ratio $r_s$, we employ the principles of similar triangles to derive the equation:
\begin{equation}
    ||P_1B||\cdot ||P_2B||=||P_1A||\cdot ||P_2C||.
\end{equation}
Here, $||P_1B||$, $||P_2B||$,$||P_1A||$, and $||P_2C||$ are line segments depicted In Fig.~\ref{fig: example_obb}. When $w\geq h$, $||P_1B||\cdot ||P_2B|| = h^2r_s(1-r_s)$. Conversely, when $w<h$, $||P_1A||\cdot ||P_2C|| = w^2r_s(1-r_s)$. Assuming $w\geq h$, we deduce:
\begin{equation}
\begin{cases}
    ||P_1B||\cdot (h-||P_1B||) = h^2r_s(1-r_s),\\
    ||P_1A||\cdot (w-||P_1A||) = h^2r_s(1-r_s).\\
\end{cases}
\end{equation}
Solving these equations yields:
\begin{equation}
\begin{cases}
    ||P_1B||=\frac{1\pm (1-2r_s)}{2}h,\\
    ||P_1A||=\frac{1\pm \sqrt{1-4\cdot\frac{h^2}{w^2}\cdot r_s(1-r_s)}}{2}w.
\end{cases}
\label{eq: sol_AB}
\end{equation}
These solutions delineate four distinct groups, corresponding to four unique OBBs. The detailed process for constructing these OBBs is elucidated in \cref{sup: decode}.

\subsection{OBB Recovery from Nine Parameters}
\label{sup: decode}
This section elucidates the method for computing the OBB from its outer HBB, sliding ratio $r_s$, and IoU scores $s_0$, $s_1$, $s_2$, and $s_3$.

Building upon the methodology outlined in \cref{sup: four_type} for computing two OBB vertices using Eq.~\ref{eq: sol_AB}, the four OBBs can be derived as follows:

\begin{equation}
    \begin{split}
        OBB_0 = &[(x_c - x_s, y_c - 0.5h),
                (x_c + 0.5w, y_c + y_s), \\
                &(x_c + x_s, y_c + 0.5h),
                (x_c - 0.5w, y_c - y_s)], \\
        OBB_1 = &[(x_c + x_s, y_c - 0.5h),
                (x_c + 0.5w, y_c + y_s), \\
                &(x_c - x_s, y_c + 0.5h),
                (x_c - 0.5w, y_c - y_s)], \\
        OBB_2 = &[(x_c - x_s, y_c - 0.5h),
                (x_c + 0.5w, y_c - y_s), \\
                &(x_c + x_s, y_c + 0.5h),
                (x_c - 0.5w, y_c + y_s)], \\
        OBB_3 = &[(x_c + x_s, y_c - 0.5h),
                (x_c + 0.5w, y_c - y_s), \\
                &(x_c - x_s, y_c + 0.5h),
                (x_c - 0.5w, y_c + y_s)]. \\
    \end{split}
    \label{eq: decoded_OBBs}
\end{equation}
Here, $y_s$ and $x_s$ are computed as:
\begin{equation}
    \begin{aligned}
        y_s &= \begin{cases}
            \frac{1-2r_s}{2}h & w\geq h,\\
            \frac{\sqrt{1-4\frac{w^2}{h^2}r_s(1-r_s)}}{2}h & w < h,
        \end{cases} \\
        x_s &= \begin{cases}
            \frac{\sqrt{1-4\frac{h^2}{w^2}r_s(1-r_s)}}{2}w & w \geq h,\\
            \frac{1-2r_s}{2}w & w < h.
        \end{cases}        
    \end{aligned}
\end{equation}
The OBB associated with the highest IoU score is selected as the recovered result.

\subsection{COBB Implementation for Oriented Proposals}
\begin{figure}[t]
\centering
\includegraphics[width=0.95\linewidth]{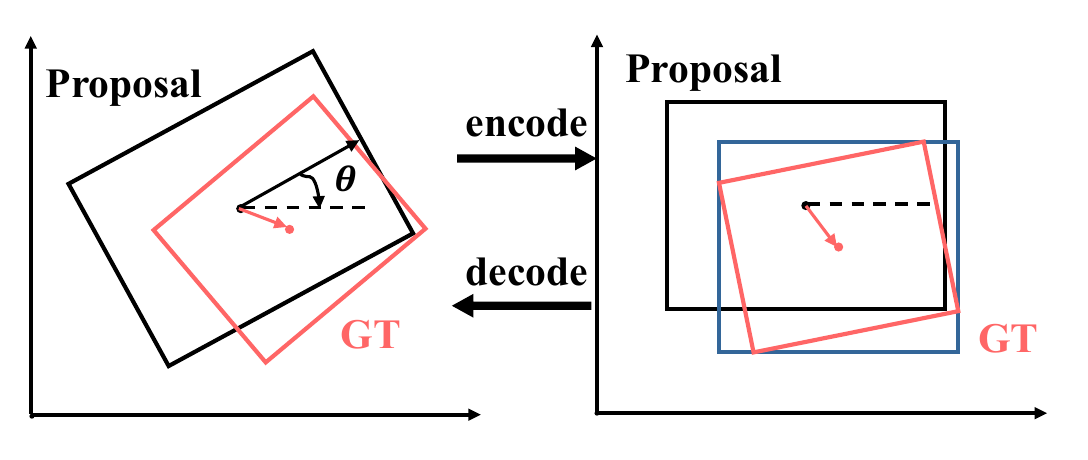}
\caption{\textbf{COBB for Oriented Proposals.}}
\label{fig: rotate}
\end{figure}

Many models utilize the discrepancy between ground truth and assigned proposals as an n-dimensional regression target. In Sec.~\ref{sec:method_target}, we introduced employing COBB to compute the bias between the ground truth OBB and a horizontal proposal region. Here, we extend this method to oriented proposal regions.

As depicted in Fig.~\ref{fig: rotate}, for an oriented proposal with center $(x_c, y_c)$ and rotation angle $\theta$ (clockwise), the regression target computation involves the following steps:
\begin{itemize}
    \item Rotate the oriented proposal and ground truth OBB by $\theta$ counterclockwise around the center $(x_c, y_c)$.
    \item Calculate the bias between the rotated proposal region and ground truth, leveraging the method introduced in Sec.~\ref{sec:method_target} for horizontally oriented regions.
\end{itemize}
For predicting the OBB from an oriented proposal and predicted vector, the process involves:
\begin{itemize}
\item Rotate the oriented proposal and ground truth OBB by $\theta$ counterclockwise around the center $(x_c, y_c)$.
\item Compute based on the rotated horizontally oriented proposal and the predicted bias to obtain an OBB.
\item Rotate the OBB calculated in the previous step by $\theta$ clockwise to derive the predicted OBB.
\end{itemize}

This methodology suits Oriented Object Detection (OOD) models using Rotated RoI Align for feature extraction from oriented regions~\cite{ding2019learning, xie2021oriented}. However, if a model utilizes a different method for feature extraction from oriented regions, the proposed regression target generation method may not be applicable.

\subsection{Relationship between $r_s$ and $r_a$}
We assert that the sliding ratio $r_s$ provides an approximation for $r_a$, the acreage ratio of the OBB concerning its outer HBB in Sec.~\ref{sec:method_details}. In this section, we establish the relationship between $r_s$ and $r_a$ and subsequently demonstrate why only a pair of symmetrical OBBs share identical $(x_c, y_c, w, h, r_a)$.

In \cref{sup: four_type}, we established that only four OBBs share the same $(x_c, y_c, w, h, r_s)$. For any OBB, we express $r_s$ in terms of $r_a$ and derive the unified equation:
\begin{equation}
    4r_s(1-r_s)=\frac{r_{wh}^2+1-\sqrt{(r_{wh}^2+1)^2-16r_{wh}^2r_a(1-r_a)}}{2r_{wh}^2}.
\label{eq: rsra}
\end{equation}

Here, $r_{wh}=\min(\frac{w}{h},\frac{h}{w})$ denotes the aspect ratio of the outer HBB. Eq.~\ref{eq: rsra} indicates that $r_s(1-r_s)$ monotonically increases with respect to $r_a(1-r_a)$. Since four OBBs share identical $x_c, y_c, w, h, r_s$, they also share the same $r_a(1-r_a)$. Given that $r_a$ for $OBB_0$ and $OBB_3$ is below $0.5$, while for the other two, it exceeds $0.5$, we conclude that only a symmetrical OBB pair shares identical $(x_c, y_c, w, h, r_a)$.

\subsection{Details of Computing IoU scores}
\label{sup: iou_scores}
Directly computing IoU scores often involves generating four OBBs based on $(x_c, y_c, w, h, r_s)$ and then evaluating the IoU between these OBBs and the ground truth. However, this direct calculation can be intricate and time-consuming. Our approach involves first computing the IoUs between the four OBBs sharing identical $(x_c, y_c, w, h, r_s)$ based on the five parameters.

The simplest method for computing IoU scores is directly generating the four OBBs according to $(x_c, y_c, w, h, r_s)$, and then calculating the IoU between the ground truth and the four OBBs. However, the calculation of IoU between OBBs is complex and time-consuming. We notice that the IoU between the four OBBs sharing the same $(x_c, y_c, w, h, r_s)$ can be directly computed from the five parameters. As a result, we can compute these IoUs first, and then select the IoU scores based on the type of the ground truth OBB.

Assuming $w\geq h$, let's define intermediate variables:
\begin{equation}    
    \begin{aligned}
        r_{sx} &= \frac{1- \sqrt{1-4\cdot\frac{h^2}{w^2}\cdot r_s(1-r_s)}}{2}, r_{sy}=r_s,\\
        l_1 &= \sqrt{(r_{sx} w)^2 + (r_{sy} h)^2}, \\
        l_2 &= \sqrt{[(1 - r_{sx}) w]^2 + [(1 - r_{sy}) h]^2}, \\
        l_3 &= \sqrt{(r_{sx} w)^2 + [(1 - r_{sy}) h]^2}, \\
        l_4 &= \sqrt{[(1 - r_{sx}) w]^2 + (r_{sy} h)^2},
    \end{aligned}
\end{equation}
where $l_1$, $l_2$, $l_3$, and $l_4$ are correspond to potential side lengths of the OBBs.

IoU between $OBB_0$ and $OBB_1$, $IoU_{0,1}$ is:
\begin{equation}
    \begin{split}
        I_{0,1} &= [1-\frac{(1-2r_{sx})r_{sx} w^2}{(1-r_{sy}) h^2}] l_1 l_2, \\
        IoU_{0,1} &= \frac{I_{0,1}}{l_1 l_2 + l_3 l_4 - I_{0,1}}.
    \end{split}
\end{equation}

IoU between $OBB_0$ and $OBB_2$, $IoU_{0,2}$ is:
\begin{equation}
    \begin{split}
        I_{0,2} &= (1-\frac{(1-2r_{sy})r_{sy}h^2}{(1-r_{sx})w^2})l_1l_2,\\
        IoU_{0,2} &= \frac{I_{0,2}}{l_1l_2+l_3l_4-I_{0,2}}.
    \end{split}
\end{equation}

IoU between $OBB_0$ and $OBB_3$, $IoU_{0,3}$ is:
\begin{equation}
    \begin{split}
        I_{0,3} &= \frac{(r_{sx}+r_{sy}-2r_{sx}r_{sy})^2}{(1-r_{sx})(1-r_{sy})}\times \frac{wh}{2}, \\
        IoU_{0,3} &= \begin{cases}
        \frac{I_{0,3}}{2l_1l_2-I_{0,3}} & I_{0,3}\neq 0,\\
        0 & I_{0,3}=0.
        \end{cases}
    \end{split}
\end{equation}

IoU between $OBB_1$ and $OBB_2$, $IoU_{1,2}$ is:
\begin{equation}
    \begin{split}
        h_1 &= \frac{1}{2}w - \frac{\frac{1}{2}-r_{sy}}{1-r_{sy}}r_{sx}w, \\
        h_2 &= \frac{1}{2}h - \frac{\frac{1}{2}-r_{sx}}{1-r_{sx}}r_{sy}h, \\
        \tan \alpha &= \frac{\frac{\frac{1}{2}-r_{sx}}{1-r_{sx}}l_4}{\frac{1}{2(1-r_{sy})}l_3}, \quad
        \tan \beta = \frac{\frac{\frac{1}{2}-r_{sy}}{1-r_{sy}}l_3}{\frac{1}{2(1-r_{sx})}l_4}, \\
        I_{1,2} &= 2\frac{\tan \alpha \tan \beta}{\tan \alpha + \tan \beta} (h_1^2+h_2^2) + 2h_1h_2, \\
        IoU_{1,2} &= 
        \begin{cases}
            \frac{I_{1,2}}{2l_3l_4-I_{1,2}} &\tan\alpha\tan\beta\neq 0,\\
            \frac{2h_1h_2}{2l_3l_4-2h_1h_2}&\tan\alpha\tan\beta=0.
        \end{cases}
    \end{split}
\end{equation}

These IoUs constitute the IoU matrix $M(w, h, r_s)$ as:
\begin{equation}
    M(w, h, r_s)=\begin{bmatrix}
        1 & IoU_{0,1} & IoU_{0,2} & IoU_{0,3}\\
        IoU_{0,1} & 1 & IoU_{1,2} & IoU_{0,2}\\
        IoU_{0,2} & IoU_{1,2} & 1 & IoU_{0,1}\\
        IoU_{0,3} & IoU_{0,2} & IoU_{0,1} & 1\\
    \end{bmatrix}.
\end{equation}
This matrix ensures continuity for $w>0, h>0, r_s\in [0, 0.5]$. Each element in row $i$ and column $j$ of $M(w, h, r_s)$ represents the IoU between $OBB_i$ and $OBB_j$.

Given a ground truth OBB, we identify its corresponding type among these four OBBs and extract the corresponding row from $M(w, h, r_s)$ as its IoU scores.

\subsection{Proof of Encoding Continuity of COBB}
In this section, we demonstrate the continuous evolution of the nine parameters in COBB as the OBB transforms.

The outer HBB and the area of an OBB change continuously during shape transformations, ensuring the continuity of $x_c$, $y_c$, $w$, $h$, and $r_a$. Eq.~\ref{eq: rsra} substantiates the continuity of $r_s$ concerning $r_a$, thereby ensuring the continuity of $r_s$ as well.

To establish the continuity of IoU scores, we consider an OBB perturbed from OBB $X$, denoted as $Y$. The similarity between $x_c$, $y_c$, $w$, $h$, and $r_s$ of $X$ and $Y$ emerges from their continuous evolution during shape transformations. Eq.~\ref{eq: decoded_OBBs} confirms the analogous construction of four OBBs using the parameters of $X$ and $Y$. IoU scores represent the overlaps between $X$ and the four OBBs sharing the same $x_c$, $y_c$, $w$, $h$, and $r_s$ as $X$. As both $X$ and the associated four OBBs remain analogous before and after perturbation, minor disturbances on $X$ do not significantly alter its IoU scores.

\begin{figure*}[t!]
    \centering
    \subfloat[\textbf{$r_s$ when Rotation}]{
        \begin{minipage}[b]{.23\textwidth}
            \centering
            \includegraphics[width=0.95\linewidth]{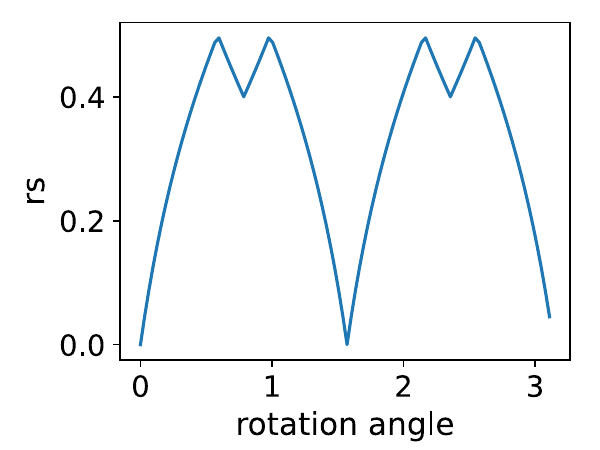}
        \end{minipage}
    }
    \subfloat[\textbf{IoU scores when Rotation}]{
        \begin{minipage}[b]{.23\textwidth}
            \centering
            \includegraphics[width=0.95\linewidth]{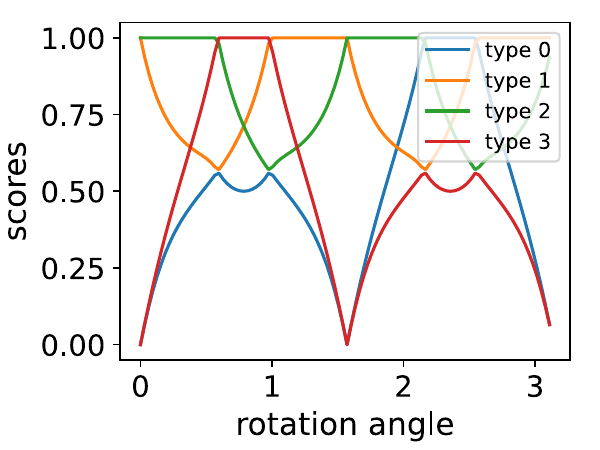}
        \end{minipage}
    }
    \subfloat[\textbf{$r_s$ when AR changes}]{
        \begin{minipage}[b]{.23\textwidth}
            \centering
            \includegraphics[width=0.95\linewidth]{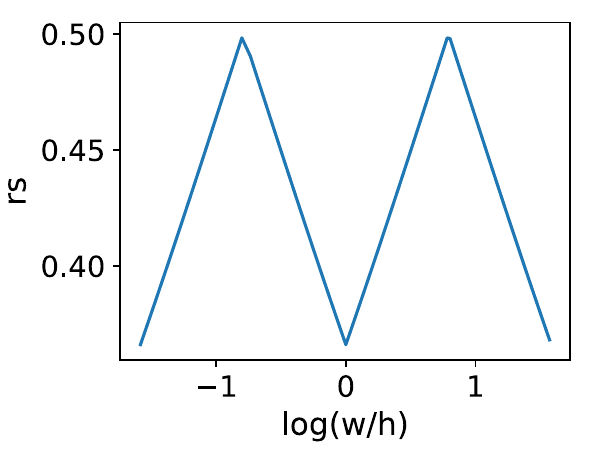}
        \end{minipage}
    }
    \subfloat[\textbf{IoU scores when AR changes}]{
        \begin{minipage}[b]{.23\textwidth}
            \centering
            \includegraphics[width=0.95\linewidth]{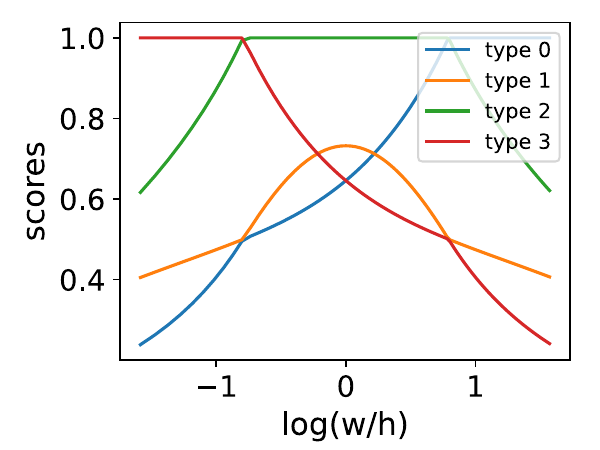}
        \end{minipage}
    }
    \vspace{-1mm}
    \caption{\textbf{COBB Parameters during OBB Transformation}. AR represents the Aspect Ratio.}
    \label{fig: curve_ours}
\end{figure*}

\begin{figure}[t!]
    \centering
    \subfloat[\textbf{$\theta$ when Rotation}]{
        \begin{minipage}[b]{.22\textwidth}
            \centering
            \includegraphics[width=0.95\linewidth]{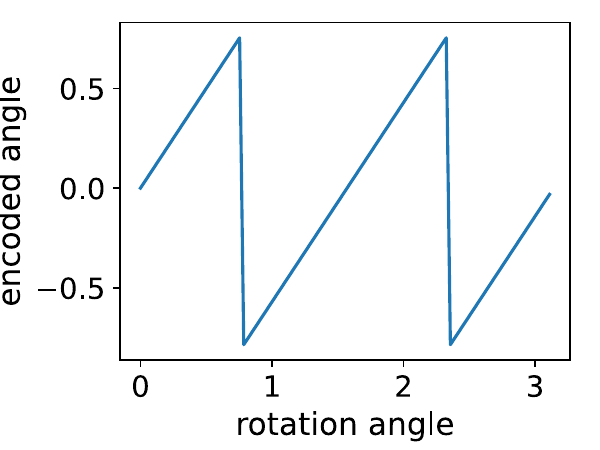}
        \end{minipage}
    }
    \subfloat[\textbf{$w$ and $h$ when Rotation}]{
        \begin{minipage}[b]{.22\textwidth}
            \centering
            \includegraphics[width=0.95\linewidth]{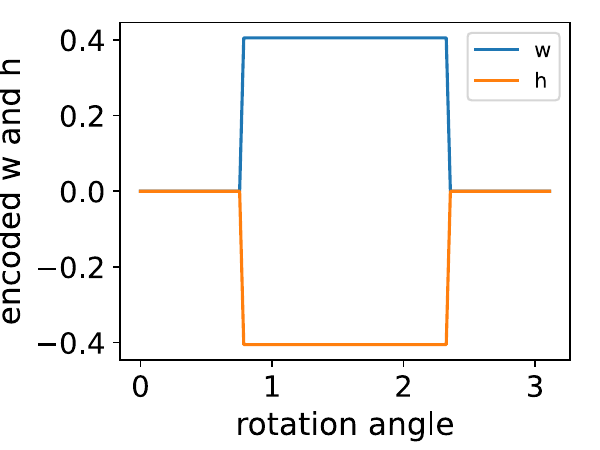}
        \end{minipage}
    }
    \vspace{-1mm}
    \caption{\textbf{Traditional Acute-angle Representation Parameters during OBB Rotation}. The traditional Acute-angle Representation encodes an OBB using the center $(x_c, y_c)$, width $w$, height $h$, and rotation angle $\theta$, which is constrained within the range $[-\frac{\pi}{4}, \frac{\pi}{4})$.}
    \label{fig: curve_acute}
\end{figure}

Fig.~\ref{fig: curve_ours} and Fig.~\ref{fig: curve_acute} provide comparative insights into the regression targets of COBB and the traditional Acute-angle Representation. COBB consistently exhibits continuous encoding results, while the Acute-angle Representation displays distinct shifts at rotation angles of $\pi/4$ and $3\pi/4$.

In summary, slight disturbances to an OBB minimally affect its $x_c$, $y_c$, $w$, $h$, $r_s$, and IoU scores, affirming the stability of these parameters under minor perturbations.

\subsection{Proof of Decoding Continuity of COBB}
Decoding Completeness is inherently fulfilled within the COBB decoding process. Here, we aim to establish the Decoding Robustness of COBB.

Slight disturbances in the parameters of COBB can be categorized as perturbations in the outer HBB, $r_s$, and IoU scores. As per Eq.~\ref{eq: decoded_OBBs}, minor perturbations in the outer HBB and $r_s$ won't significantly alter the decoded OBB if IoU scores remain constant. If a perturbation doesn't affect the classification corresponding to the highest IoU score, it won't impact the decoded OBB. Even if a perturbation shifts the highest score from classification $i$ to $j$, where $s_i = 1$ (indicating the correct classification), slight disturbances maintain the relative values of $s_i$ and $s_j$, making $s_j$ close to 1. As $s_j$ represents the IoU between the correct OBB and the decoded $OBB_j$ (from Eq.~\ref{eq: decoded_OBBs}), choosing $OBB_j$ as the decoded result won't introduce significant decoding errors.

\subsection{Comparison of Different Regression Targets}
To mitigate the sensitivity of objects with large aspect ratios to predicted results, we introduce $r_{ln}$ as the regression target, which is defined by 
Eq.~\ref{eq:rln}. It is an approximation to $f_{ln}(r_a)=1+\log_2(r_a)$. The direct use of $f_{ln}(r_a)$ is feasible by recovering $r_s$ with Eq.~\ref{eq: rsra}. Notably, only two OBBs share the same $x_c$, $y_c$, $w$, $h$, and $r_{ln}$; hence, we need to derive $r_s$ from $r_{ln}$ first and then select one of the four OBBs that share the same $r_s$ as the decoded result, preventing Decoding Ambiguity.

In previous experiments in Tab.~\ref{tab:r_target}, we established the superiority of $r_{ln}$ over $f_{ln}(r_a)$. One reason is that the process of recovering $r_s$ from $r_a$ can introduce precision errors. This section presents another reason: the sensitivity of the decoded OBB to slight disturbances in $f_{ln}(r_a)$.

\begin{figure}[t!]
    \centering
    \begin{minipage}[b]{.45\textwidth}
        \centering
        \includegraphics[width=0.95\linewidth]{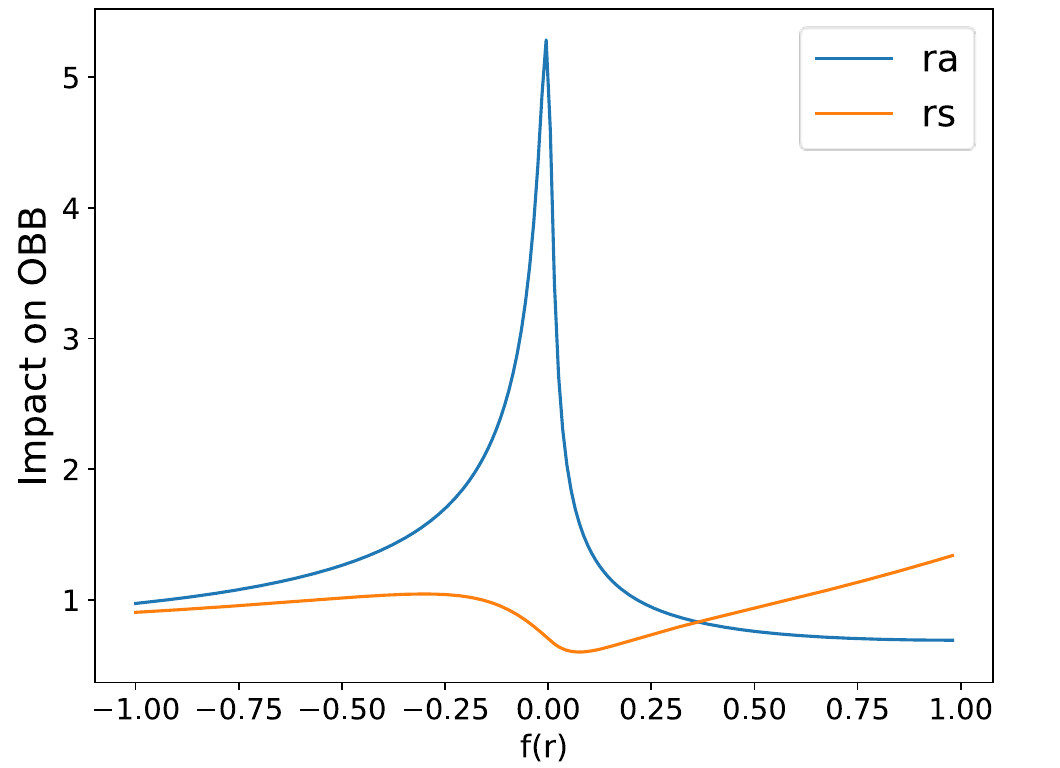}
    \end{minipage}%
    \vspace{-7pt}
    \caption{\textbf{Impact of Slight Disturbance on $r_s$ and $r_a$:} We assessed the effects on the decoded OBB when $r_s$ and $r_a$ undergo slight disturbances. To mitigate the sensitivity of OBBs with large aspect ratios to small perturbations in $r_s$ or $r_a$, we consider $f(r_s)=r_{ln}$ (as previously defined) and $f_{ln}(r_a)=1+\log_2r_a$ as the specific perturbed values.}
    \label{fig: rsra}
\end{figure}
\begin{table}[t!]
	\renewcommand\arraystretch{1.4}
	\setlength\tabcolsep{4pt}
	\footnotesize
	\centering		
    \caption{\textbf{NAE of parameters.} \textbf{$\alpha$} is the parameter for orientation determination in Gliding Vertex. In our method and Gliding Vertex \textbf{xy} denotes the center, \textbf{wh} represents width and height, \textbf{$r_s$} and \textbf{IoU scores} are parameters of our method, and \textbf{$\alpha$} denotes the additional parameters for orientation determination of Gliding Vertex.}
    \label{tab:NAE}
	\vspace{-3mm}
    \resizebox{0.42\textwidth}{!}
{
		\begin{tabular}{cccccc}
\toprule
\textbf{Method} &  \textbf{xy} & \textbf{wh} & \textbf{$r_s$} & \textbf{IoU scores} & \textbf{$\alpha$} \\
\hline
Ours. & 0.18\% & 0.31\% & 0.28\% & 0.41\% & -\\
Glidng Vertex & 0.17\% & 0.31\% & - & - & 0.75\%\\
\bottomrule
		\end{tabular}
 }
 	\vspace{-3mm}
 \end{table}

\begin{table*}[t!]
	\renewcommand\arraystretch{1.4}
	\setlength\tabcolsep{4pt}
	\footnotesize
	\centering		
    \caption{\textbf{$\text{mAP}_{50}$ of Models in JDet Benchmark on DOTA-v1.0.}}
    \label{tab: detailed_benchmark}
	\vspace{-3mm}
    \resizebox{0.98\textwidth}{!}
{
		\begin{tabular}{ccccccccccccccccccc}
\toprule
		\textbf{Model} & \textbf{PL} & \textbf{BD} & \textbf{BR} & \textbf{GTF} & \textbf{SV} & \textbf{LV} & \textbf{SH} & \textbf{TC} & \textbf{BC} & \textbf{ST} & \textbf{SBF} & \textbf{RA} & \textbf{HA} & \textbf{SP} & \textbf{HC} & \textbf{$\text{mAP}_{50}$}\\ \hline
CSL & 89.36 & 77.74 & 37.66 & 66.73 & 77.85 & 59.27 & 77.75 & 90.86 & 79.74 & 82.23 & 55.20 & 63.07 & 51.55 & 66.03 & 44.93 & 67.99\\
RsDet & 89.47 & 78.02 & 38.39 & 63.01 & 77.28 & 60.95 & 76.41 & 90.77 & 81.80 & 83.47 & 54.26 & 61.98 & 51.68 & 66.55 & 52.18 & 68.41\\
RetinaNet & 89.56 & 79.70 & 38.12 & 67.58 & 76.21 & 59.47 & 76.80 & 90.70 & 83.56 & 81.07 & 53.88 & 64.73 & 53.36 & 65.99 & 41.97 & 68.18\\
RetinaNet+KLD & 89.42 & 76.51 & 39.36 & 65.20 & 77.83 & 63.19 & 82.48 & 90.90 & 79.20 & 83.46 & 54.39 & 63.83 & 53.26 & 67.67 & 44.53 & 68.75\\
RetinaNet+KFIoU & 89.38 & 81.05 & 39.12 & 68.39 & 77.37 & 62.58 & 77.86 & 90.87 & 82.43 & 82.32 & 56.00 & 65.60 & 53.50 & 67.08 & 41.36 & 68.99\\
RetinaNet+GWD & 89.10 & 78.04 & 39.07 & 69.21 & 77.27 & 62.05 & 81.05 & \textbf{90.90} & 84.31 & 83.02 & 57.24 & 61.96 & 53.79 & 64.40 & 43.83 & 69.02\\
FCOS & 88.07 & 78.80 & 44.85 & 65.58 & 74.88 & 68.17 & 77.84 & 90.90 & 79.99 & 83.60 & 57.42 & 65.30 & 62.96 & 68.59 & 48.55 & 70.37\\
ATSS & 88.44 & 78.77 & 49.14 & 67.17 & 77.63 & 74.97 & 85.56 & 90.78 & 83.39 & 83.82 & 59.17 & 61.39 & 65.35 & 65.48 & 55.50 & 72.44\\
$\text{S}^2\text{A-Net}$ & 89.21 & 81.04 & 50.97 & 71.35 & 78.21 & 77.42 & 87.05 & 90.88 & 82.51 & 85.00 & 63.35 & 64.52 & 66.45 & 67.67 & 53.60 & 73.95\\
Faster R-CNN & 89.46 & 83.89 & 49.64 & 69.59 & 77.57 & 73.23 & 86.52 & 90.90 & 79.33 & 85.74 & 58.84 & 60.49 & 65.78 & 68.64 & 55.55 & 73.01\\
Gliding. & 89.34 & 83.68 & 50.15 & 69.97 & 78.20 & 72.51 & 87.17 & 90.90 & 79.94 & 85.46 & 57.07 & 62.57 & 66.93 & 66.12 & 59.61 & 73.31\\
RoI Trans. & 89.21 & 83.88 & 53.01 & \textbf{72.97} & 77.86 & 78.08 & 88.01 & 90.86 & 86.94 & 85.84 & 63.50 & 61.53 & 75.77 & \textbf{70.33} & 56.02 & 75.59\\
O-RCNN. & \textbf{89.72} & 84.41 & 52.94 & 71.80 & \textbf{78.71} & 77.51 & \textbf{88.15} & 90.90 & 85.90 & 84.90 & 61.58 & 63.93 & 74.23 & 69.94 & 51.99 & 75.11\\
ReDet & 88.90 & 82.28 & 52.42 & 72.76 & 77.63 & 82.52 & 88.12 & 90.88 & 86.51 & 85.81 & \textbf{67.34} & 65.33 & 76.07 & 68.58 & 60.54 & 76.38\\
\rowcolor{gray!20} Ours (RoI Trans. based) & 89.52 & \textbf{84.98} & \textbf{54.99} & 72.16 & 77.71 & 82.81 & 88.10 & 90.81 & 85.45 & 85.62 & 63.89 & \textbf{66.15} & \textbf{76.64} & 70.13 & 59.05 & \textbf{76.53}\\
\rowcolor{gray!20} Ours (ReDet based) & 89.71 & 84.82 & 53.27 & 71.40 & 77.02 & \textbf{83.80} & 88.07 & 90.85 & \textbf{87.11} & \textbf{86.20} & 66.44 & 63.72 & 76.15 & 67.99 & \textbf{61.26} & 76.52\\
\bottomrule

		\end{tabular}
 } 
 \end{table*}

\begin{table}[t!]
	\renewcommand\arraystretch{1.4}
	\setlength\tabcolsep{4pt}
	\footnotesize
	\centering		
    \caption{\textbf{$\text{mAP}$ of COBB on DOTA-v1.0 under Multi-scale Data Augmentation.}}
    \label{tab: cobb_multiscale}
	\vspace{-3mm}
    \resizebox{0.4\textwidth}{!}
{
		\begin{tabular}{ccccc}
\toprule
 \textbf{Models} & \textbf{$\text{mAP}_{50}$} & \textbf{$\text{mAP}_{75}$} & \textbf{$\text{mAP}_{50:95}$} \\
\hline
Oriented R-CNN~\cite{xie2021oriented} & 78.73 & 55.07 & 50.57\\
\rowcolor{gray!20} +COBB-sig & 79.09 & 55.61 & \textbf{50.80}\\
\rowcolor{gray!20} +COBB-ln & \textbf{79.23} & \textbf{56.15} & 50.55\\
\bottomrule
		\end{tabular}
 } 
 \end{table}

 \begin{table}[t!]
	\renewcommand\arraystretch{1.4}
	\setlength\tabcolsep{4pt}
	\footnotesize
	\centering		
    \caption{\textbf{Experiments on Optimizing IoU Enhancement Factor $\lambda$.} We utilize $IoU^\lambda$ as a specific regression target for IoU scores to widen the distinction between scores for the ground truth and other classifications. These experiments were conducted using Rotated Faster R-CNN + COBB-sig.}
    \label{tab: IoU_scores}
	\vspace{-3mm}
    \resizebox{0.28\textwidth}{!}
{
		\begin{tabular}{cccc}
\toprule
 \textbf{$\lambda$} & \textbf{$\text{mAP}_{50}$} & \textbf{$\text{mAP}_{75}$} & \textbf{$\text{mAP}_{50:95}$} \\
\hline
1 & 73.41 & \textbf{44.21} & 43.10\\
2 & \textbf{74.00} & 44.03 & \textbf{43.29}\\
4 & 73.63 & 43.85 & 42.83\\
\bottomrule
		\end{tabular}
 } 
 \end{table}

 \begin{table}[t!]
	\renewcommand\arraystretch{1.4}
	\setlength\tabcolsep{4pt}
	\footnotesize
	\centering		
    \caption{\textbf{Experiments on SOTA baselines on DOTA-v1.0.}}
    \label{tab:SOTA}
	\vspace{-3mm}
    \resizebox{0.42\textwidth}{!}
{
		\begin{tabular}{llll}
\toprule
\textbf{Method} &  \textbf{$\text{mAP}_{50}$} & \textbf{$\text{mAP}_{75}$} & \textbf{$\text{mAP}_{50:95}$} \\
\hline
SES & 75.72 & 48.86 & 46.19\\
SES + Ours. & 76.43(\textcolor{red}{+0.71}) & 49.28(\textcolor{red}{+0.42}) & 46.59(\textcolor{red}{+0.40})\\\hline
LSKNet-t & 76.68 & 49.28 & 46.15 \\
LSKNet-t + Ours. & 77.29(\textcolor{red}{+0.61}) & 50.91(\textcolor{red}{+1.63}) & 47.62(\textcolor{red}{+1.47}) \\
\bottomrule
		\end{tabular}
 }
 	\vspace{-5mm}
 \end{table}
Fig.~\ref{fig: rsra} illustrates the impact of slight disturbances on $r_{ln}$ and $f_{ln}(r_a)$. The impact is quantified as follows:
\begin{equation}
    \Delta OBB=(1-IoU[f_{Dec}(r), f_{Dec}(r+\epsilon)])\epsilon^{-1}.
\end{equation}

Here, $r$ is $r_{ln}$ or $f_{ln}(r_a)$, $f_{Dec}$ decodes $r$ with a nearly square outer HBB into an OBB, and $\epsilon$ is a small value close to 0. Notably, when $f_{ln}(r_a)$ approaches zero, the decoded OBB demonstrates significant sensitivity to variations in $f_{ln}(r_a)$, whereas the impact of slight disturbances on $r_{ln}$ remains relatively stable.

\subsection{Models' Ability to Well Estimate Parameters}

We contend that the parameters in our COBB are easily estimable due to their continuity. To elucidate this assertion, we introduce the Normalized Average Error (NAE) as a metric for assessing the difficulty of parameter estimation. Given the $i$-th prediction of the parameter as $P_i$ and its corresponding ground truth as $T_i$, the NAE is defined as $\textrm{NAE}=\frac{1}{N}\sum_{i=1}^N\frac{(P_i - T_i)^2}{(\max(T) - \min(T))^2}$, where $N$ represents the number of predictions, and $\max(T)$ and $\min(T)$ denote the maximum and minimum values of ground truth values.

We posit that parameters with small NAE values are more ready to estimate. The NAE values of our method and Gliding Vertex are documented in Tab.~\ref{tab:NAE}. Without significantly influencing the prediction difficulty of the other parameters, $r_s$ and IoU scores in our method can be better estimated than $\alpha$s in Gliding Vertex.

\section{Additional Experiments}
\subsection{Training Settings}
Experiments were conducted using Jittor~\cite{hu2020jittor} on a single RTX 3090 running on Linux. The models utilized ResNet-50~\cite{he2016deep} with FPN~\cite{lin2017feature} to extract multi-level feature maps. During training, an SGD optimizer was employed, with a learning rate of 0.005 for two-stage models and 0.01 for one-stage models.

For dataset-specific training:

\begin{itemize}
    \item DOTA, FAIR1M, and DIOR datasets were trained for 12 epochs, while HRSC2016 was trained for 36 epochs.
    \item For images in the DOTA and FAIR1M datasets larger than $1,024\times 1,024$, they were split into multiple $1,024\times 1,024$ tiles with a 200-pixel overlap.
    \item Data augmentation included random horizontal and vertical flips, each with a $50\%$ probability.
\end{itemize}

\subsection{More Results on JDet Benchmark}
Tab.~\ref{tab: detailed_benchmark} provides a detailed breakdown of the results presented in Tab.~\ref{tab:base}. These comprehensive experimental findings underscore the superior performance achieved by our proposed method within our benchmark evaluation.

Analysis of the results indicates a pronounced advantage of our method in detecting objects characterized by a long aspect ratio, exemplified by categories such as Bridge (BR), Large Vehicle (LV), and Harbor (HA). This advantage is attributable to the inherent continuity embedded within our methodology, which mitigates potential confusion and interference during the training process arising from sudden changes in regression targets as OBBs approach a horizontal orientation. Notably, the models implemented in our benchmark employ the Acute-angle Representation as a default approach. The conspicuous discontinuity inherent in the Acute-angle Representation becomes especially evident when objects exhibit a considerable aspect ratio. Consequently, the discernible advantage exhibited by our proposed method in these scenarios underscores the efficacy of its continuous nature.

\subsection{Futher Ablation Study for COBB}

\textbf{COBB Performance under Multi-scale Data Augmentation.} Tab.~\ref{tab: cobb_multiscale} showcases the performance of COBB on DOTA-v1.0 under Multi-scale data augmentation. Multi-scale augmentation involved resizing training images to 0.5, 1.0, and 1.5 times their original dimensions, and these variations were incorporated into training and testing. For large images split into tiles, the width of the overlapping area was adjusted to 500 pixels. The results affirm the efficacy of COBB under this data augmentation technique.

\textbf{Optimizing the IoU Enhancement Factor $\lambda$.} We alter the IoU scores by exponentiating them to the power of $\lambda$ to diminish the impact of incorrect categories. Tab.~\ref{tab: IoU_scores} displays experiments demonstrating that $\lambda=2$ achieves optimal performance. Notably, $\lambda=1$ exhibits advantages in $\text{mAP}_{75}$. Throughout this article, COBB-sig employs $\lambda=2$, while COBB-ln uses $\lambda=1$.

\subsection{Experiments on Latest Techniques}
To verify the advantage of our method on SOTA methods, we added our method to the latest SOTAs, including SES~\cite{yang2023sampling} and LSKNet~\cite{li2023large}. Experiments in Tab.~\ref{tab:SOTA} demonstrate that our method achieves a relatively large $\text{mAP}_{50}$ improvement, 0.71\% and 0.61\%, over SES and LSKNet, respectively.

\end{document}